\documentclass[11pt]{article}
\usepackage{UF_FRED_paper_style}

\usepackage{xspace}
\usepackage{booktabs}
\usepackage{graphicx}
\usepackage{amsmath}
\usepackage{subcaption}
\usepackage{colortbl}
\usepackage{pgfplots}
\usepackage{array}
\newcolumntype{P}[1]{>{\centering\arraybackslash}p{#1}}
\usepackage{tcolorbox}
\usepackage{soul}
\usepackage{placeins}
\usepackage{lineno}

\newcommand{\gptfour}{\textsc{GPT-4}\xspace}
\newcommand{\gptfive}{\textsc{GPT-5}\xspace}
\newcommand{\llm}{\textsc{LLM}}
\newcommand{\phq}{\textsc{PHQ-9}\xspace}

\newcommand{\hlc}[2][yellow]{{%
    \colorlet{foo}{#1}%
    \sethlcolor{foo}\hl{#2}}%
}

\usepackage{lipsum}  

\title{Explaining GPTs' Schema of Depression: A Machine Behavior Analysis}

\author{
Adithya V Ganesan$^{1*}$ \and 
Vasudha Varadarajan$^{2}$ \and 
Yash Kumar Lal$^{1}$ \and 
Veerle C. Eijsbroek$^{3}$ \and 
Katarina Kjell$^{3}$ \and 
Oscar N.E. Kjell$^{1,3}$ \and 
Tanuja Dhanasekaran$^{4}$ \and 
Elizabeth C Stade$^{5,6}$ \and 
Johannes C Eichstaedt$^{5,6}$ \and
Ryan L Boyd$^{7}$ \and 
H Andrew Schwartz$^{1,8}$ \and 
Lucie Flek$^{9,10}$ \and 
\\
{\small
\begin{tabular}{@{}c@{}}
$^{1}$Stony Brook University \;
$^{2}$Carnegie Mellon University \;
$^{3}$Lund University \\
$^{4}$Independent Researcher \;
$^{5}$Stanford University \\
$^{6}$Institute of Human-Centered AI at Stanford University \;
$^{7}$The University of Texas at Dallas \\
$^{8}$Vanderbilt University \;
$^{9}$University of Bonn \\
$^{10}$The Lamarr Institute for Machine Learning \& Artificial Intelligence
\end{tabular}
} \\ \\
$^{*}$Corresponding Author: \texttt{avirinchipur@cs.stonybrook.edu}
}

\begin{document}

\maketitle
\begin{abstract}

Use of large language models such as ChatGPT (\gptfour/\gptfive) for mental health support has grown rapidly, emerging as a promising route to assess and help people with mood disorders, like depression. 
However, we have a limited understanding of these language models' schema of mental disorders, that is, how it internally associates and interprets symptoms of such disorders. 
In this work, we leveraged contemporary measurement theory to decode how \gptfour and \gptfive interrelates depressive symptoms, providing an explanation of how LLMs apply what they learn and informing clinical applications. 
We found that \gptfour: (a) had strong convergent validity with standard instruments and expert judgments $(r=.70-.81)$, and 
(b) behaviorally linked depression symptoms with each other (symptom inter-correlates $r=.23-.78$)  in accordance with established literature on depression; however, it 
(c) underemphasized the relationship between \textit{suicidality} and other symptoms while overemphasizing \textit{psychomotor symptoms}; and 
(d) suggested novel hypotheses of symptom mechanisms, for instance, indicating that \textit{sleep} and \textit{fatigue} are broadly influenced by other depressive symptoms, while \textit{worthlessness/guilt} is only tied to \textit{depressed mood}.
\gptfive showed a slightly lower convergence with self-report, a difference our machine-behavior analysis makes interpretable through shifts in symptom–symptom relationships.
These insights provide an empirical foundation for understanding language model's mental health assessments and demonstrate a generalizable approach for explainability in other models and disorders. 
Our findings can guide key stakeholders to make informed decisions for effectively situating these technologies in the care system.\\

\noindent

\end{abstract}

\newpage
\section{Introduction}
Large Language Models (LLMs), such as OpenAI's \gptfour or Google's Gemini are starting to be actively used by thousands of individuals~\citep{hua2025scoping} for mental health support, although we have a limited understanding of their outcomes and limitations. 
A recent survey in the \textit{Journal of Medical Economics} revealed that over 1 in 10 clinicians in the USA use ChatGPT, and roughly 50\% are interested in continuing its use for data entry, medical scheduling, or research~\citep{shryock-2023-AI}.
However, similar to treating mental disorders with drugs that are not understood, there are many risks in using LLMs for mental health support without explanation of how they behave in the domain~\citep{stade2024large, derga-etal-2024-ChatGPT}.

This paper presents a framework for evaluating and explaining LLMs' ``representation'' of mental health. 
To do this, we apply principles from measurement theory~\cite{rust2021modern} to infer \gptfour's and \gptfive's \textit{schema} of depression (how it appears to organize, interpret, and relate depressive symptoms) based on its behavior.
In the process, we quantitatively evaluate two fundamental questions: (a) how does GPT identify and estimate depression symptoms from language, and (b) what is its latent structure between the symptoms used to form an overall assessment.

We utilize this framework to study GPTs' structural representation of depression for several reasons. 
First, depression is one of the most prevalent mental health conditions, affecting an estimated 10–15\% of the population throughout their lifetime~\cite{lepine2011increasing, lim2018prevalence}, and is the leading cause of disability worldwide~\cite{friedrich2017depression}. 
Second, there exists a vast body of research on depression's symptoms, structure, and their inter-relationships, providing a strong theoretical reference against which to compare~\cite{watson2009differentiating, waszczuk2017hierarchical, fried2016good}. 
Lastly, depression is a highly heterogeneous disorder comprised of cognitive, affective, behavioral, and somatic symptoms~\cite{fried2015depression}; depression is also highly co-morbid with other conditions, including anxiety~\cite{leckman1983panic}, substance use~\cite{swendsen2000comorbidity}, somatic disorders~\cite{katon1982depression} and other internalizing conditions~\cite{kessler2005prevalence}.
Thus, understanding \gptfour's schema of depression suggests generalizations to many aspects and forms of psychopathology.

While previous work on \textit{how} GPTs' assessments are formed is limited, there is growing evidence of these models' ability to predict mental disorder~\cite{shin2024using} as compared to self-report questionnaires~\cite{galatzerlevy2023capability} or experts~\cite{levkovich2025evaluating}. 
Such methods reveal \textit{what} the model can do but do not explain \textit{how} (i.e. explainability) it performs such tasks. 
The responsible use of LLMs, particularly in critical applications that involve human health and well-being, requires explaining how they perform their tasks.

To address the complexities of explaining a massive model, we use a \textit{machine behavior} approach~\cite{rahwan2019machine}.  
Over the years, behavioral science has developed and continually refined tools and methods to explain (measure, interpret, and understand) the human mind using mathematical and empirical methods through behavioral observation~\cite{galatzer-levy-etal-2023-machine}. 
Such methods have uncovered \textit{schemas} --- the internal concepts underlying a process --- embedded within the neural system from observations of human behaviors~\cite{stein1992schemas, schmidt1995schema, waters2006attachment}.
Drawing parallels from cognitive and behavioral psychology, our experiments are designed to collect extensive observations of \gptfour's ability to assess depression. 
We analyze its generated text (responses) and the input prompts (stimuli) to infer \gptfour's and \gptfive's schema of depression. 
This inferred schema is used to explain the models' behavior in assessing depression. 

\begin{figure*}[!t]
    \centering
    \caption{\textbf{Overview of Machine Behavior Evaluation} to uncover the structure of depression in \gptfour.
    (1) \gptfour is prompted with step-by-step instructions on performing depression assessment on human-written essays.
    (2) \gptfour carried out assessments on data from Gu et al., 2024~\cite{gu2024natural}, containing depression essays written by individuals. 
    The model estimated the severity scores for \phq symptoms explicitly mentioned in the essay, followed by scoring the remaining symptoms that were implicit in the essay. 
    (3) \gptfour's estimated scores and spans are used to infer its internal structure of depressive symptoms, which revealed how the web of symptoms was connected, and what language \gptfour reliably used to estimate depression severity.   
    }
    \includegraphics[width=\linewidth]{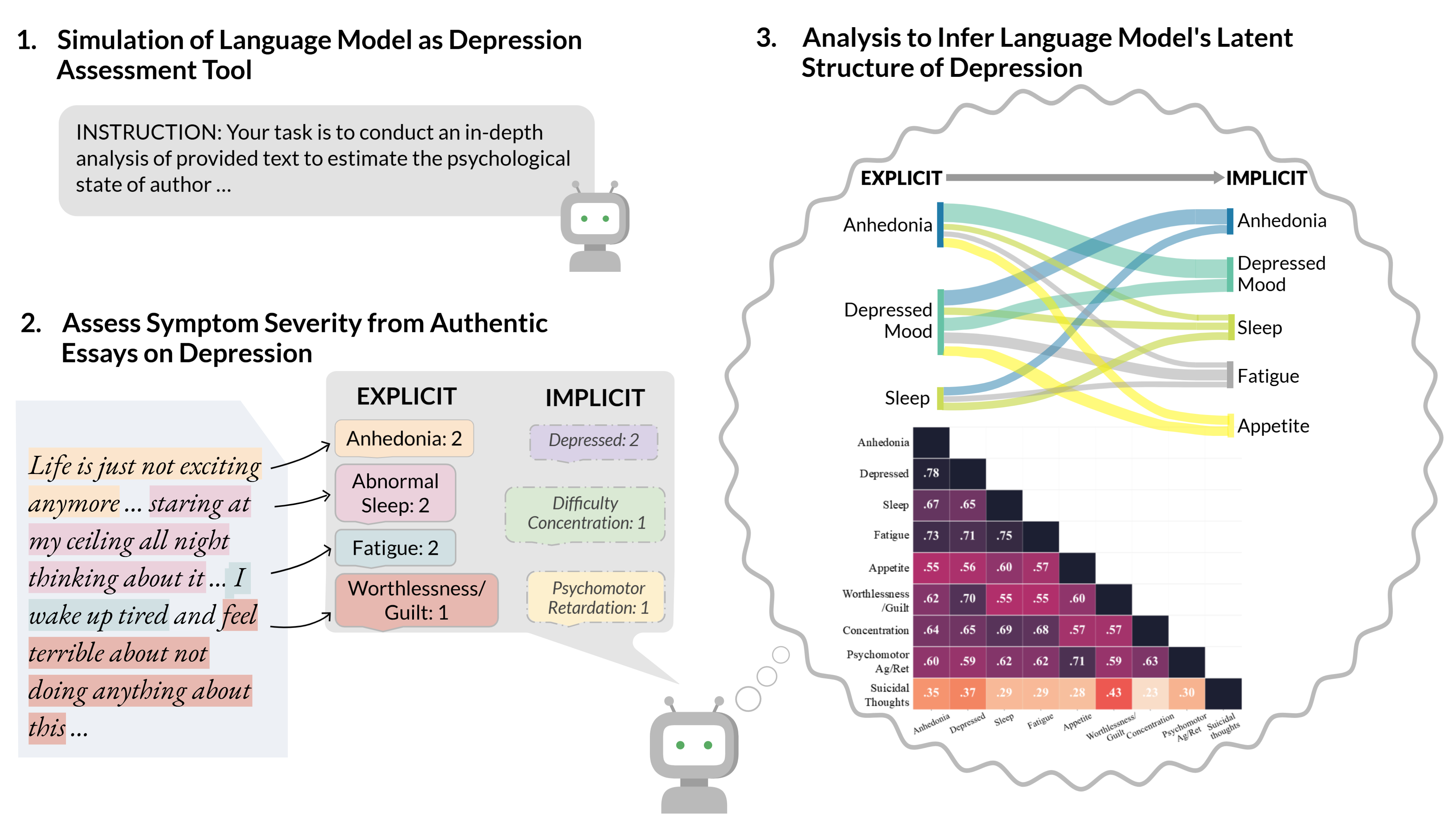}
    \label{fig:pilot}
\end{figure*}

Rahwan et al., (2019)~\cite{rahwan2019machine} and
Markowitz \& Boyd (2024)~\citep{markowitz_silicon_2024} call for such an approach to explain GPTs' functioning, and to the best of our knowledge, our study represents the first. 
We employ a robust methodological framework that can be extended to other LLMs to understand how psychopathology is represented within them.
This is depicted in \autoref{fig:pilot} and has been described extensively under Methods. 

The unique contributions of this study center on explaining \gptfour's and \gptfive's behavior when identifying depression risk within open-ended essays. 
In a sample of 955 individuals, we employ measurement theory to infer GPTs' schema of depression by analyzing (a) the relationships among key depression symptoms and (b) the role of each symptom in determining the degree of depression within the language model. 
This inferred schema is compared with one derived from self-reported scores to quantitatively identify similarities and differences in the conceptualization of depression as a network of interconnected symptoms. 
We applied differential language analysis~\cite[DLA]{schwartz2013personality} to identify the linguistic cues associated with \gptfour's identification of depressive risk factors.
Following this, we evaluated whether \gptfour shows evidence that it uses the phrases it cited as explanations for producing its depression scores.
Lastly, we used multivariate linear regression to assess how \gptfour integrates explicitly mentioned factors in essays to estimate risks for unmentioned factors, to uncover its mechanism behind symptom estimation.
We further apply this framework to \gptfive and show that performance differences relative to \gptfour on depression assessment are accounted for by shifts in the learned schema, linking aggregate assessment scores to interpretable internal relations among symptoms.

\section{Methods}\label{sec:methods}

\subsection{Data} 
\noindent\textbf{Collection.} 
Data comes from study participants that were recruited through Prolific\cite{palan2018prolific}, an online platform.
From 1040 participants in total, 258 (24.8\%) screened participants reported having been diagnosed with Major Depressive Disorder (MDD), 259 (24.8\%) with Generalized Anxiety Disorder (GAD), and 491 (47.2\%) participants were recruited from an unscreened population. 
32 (3.2\%) respondents were excluded for not answering the control items correctly. 
Removing participants that failed attention check and incomplete survey responses, the dataset comprised 963 participants, 571 identified as female, 388 as male, 2 preferred not to say, and 2 did not respond. 
The average age was 33.3 (SD = 11.2; range = 18 – 77) years. 
Participants reported being from the U.S. (n = 284) and the U.K. (n = 679).

The Swedish National Ethics Review Board deemed the research study (Ethics Application 2020-00730) exempt from requiring ethical approval according to Swedish Law (see §§ 3-4 of the Act [2003:460] on ethical review of research involving humans in Sweden). 
The research was carried out according to the Declaration of Helsinki. 
The sample comprised a mixture of screened and unscreened respondents.
More details on data can be found in \citeauthor{gu2024natural}, \citeyear{gu2024natural}.\\

\noindent\textbf{Measures.} Below we only describe the measures used in this study – for all details regarding the data please refer to Gu et al., 2023~\cite{gu2024natural}. 
Participants were asked to describe their depression using open-ended language response.
The questions was adapted from previous research\cite{kjell2019semantic} and validated in Gu et al., 2024~\cite{gu2024natural}.
Responses for the following prompt were used in this work.

\begin{tcolorbox}[
    width=\textwidth,
    colback=white,
    colframe=black,
    arc=4mm,
    boxrule=0.5pt,
    left=2mm,
    right=2mm,
    top=2mm,
    bottom=2mm,
    fonttitle=\bfseries,
    ]

\begin{tcolorbox}[
    colback=gray!5,
    boxrule=0pt,
    colframe=white,
    left=0pt,
    right=0pt,
    top=0pt,
    bottom=0pt,
    ]

\small

Over the last 2 weeks, have you been depressed or not? 
Please answer the question by typing at least a paragraph below that indicates whether you have been depressed or not. Try to weigh the strength and the number of aspects that describe if you have been depressed or not so that they reflect your overall personal state of depression. For example, if you have been depressed, then write more about aspects describing this, and if you have not been depressed, then write more about aspects describing that. 
Write about those aspects that are most important and meaningful to you. 
Write at least one paragraph in the box.

\end{tcolorbox}
\end{tcolorbox}

Following this, participants were administered \phq\cite{kroenke2001phq}, a 9 item questionnaire used to assess symptoms of major depressive disorder as defined in the DSM-IV~\cite{frances1995dsm}. 
The rating scale included nine items targeting symptoms such as “Feeling down, depressed, or hopeless”, referring to experiences from the last two weeks. 
A four-point rating scale was used (0 = Not at all to 3 = Nearly every day) to indicate the experienced severity of each symptom.\\

\noindent\textbf{Procedure. } Participants were first given information about the study, including information that participation was voluntary, that they could withdraw at any time, and that their answers were anonymous. 
Before starting the study, all participants consented to take part in the study and to share their anonymized data openly. 
The survey had three parts: First, participants answered open-ended questions concerning depression and anxiety/worry, which were presented in random order across participants. 
Second, they filled out the rating scales, which were randomized. 
Last, they reported the demographics, and mental health-related questions like sick leave and healthcare visits. 
The median time to complete the study was 20.1 minutes (mean=24.0, SD=12.8).

\subsection{Experiment Design} 
\noindent\textbf{Prompt. } Text response describing depression was sent to \gptfour\footnote{versioned 11/06 and referred to as gpt4-1106 in OpenAI's GPT directory} models along with the instructions to carry out assessment (made available as supplementary materials). 
The instruction was composed into three parts: 
\begin{enumerate}
    \item \textbf{Task description:} A short paragraph describing the task of estimating the \phq item scores by analyzing the human-written text. 
    The description included the commonly abbreviated form of the 9 items from the \phq questionnaire. 
    \item \textbf{Task Steps:} A detailed step-by-step instructions dictating \gptfour to first estimate the scores for the items that are explicitly stated in the text, followed by scoring the items that were not explicitly stated and finally, concluding with the combined PHQ9 score and severity category.
    \item  \textbf{Output Format:} \gptfour was instructed to format its generation into a predefined data structure (JSON object) to facilitate easy processing of data for analysis. 
\end{enumerate}
Although, \gptfour generated outputs fulfilled the completion of \phq estimation for all the data, it was not completely faithful to the details of the instructions. 
There were a few issues observed with following the steps detailed in the instructions, especially step 1. 
59 (6.1\%) out of 956 samples, had small issues that were manually rectified. These included cases where \gptfour had \begin{enumerate}
\item made subtle changes to the symptom abbreviations (E.g. 'Feelings of Worthlessness or Guilt' instead of 'Worthlessness or Guilt'), 
\item identified specific aspects of the symptoms (E.g. 'Insomnia' in 'Insomnia or Hypersomnia', 'Psychomotor Retardation' in 'Psychomotor Agitation/Retardation', 'Poor Appetite' in 'Poor Appetite or overeating' etc.), 
\item split the individual steps into separate JSONs, instead of nesting them within a single JSON as described in the prompt. 
\item included a summary of the assessment which was a short essay of its JSON output
\item included symptoms that were not a part of the \phq (E.g. Financial Worries, Social Exclusion)
\end{enumerate}
For the first two cases, the symptoms were manually corrected to match the abbreviations in Step 1 of the instructions. 
Type 3 generations were rectified by manually reformatting, and the summaries in type 4 were removed.
The symptoms in type 5 generations were left out of the analysis. 

\gptfour's generations varied based on its hyperparameters. 
For this work, the temperature value was set to 0.0 to select the most likely token at each step and to allow reproducibility of experiments. 
Other hyperparameters include frequency penalty (0.1), presence penalty (0.0), top p (1.0), and max tokens (450). \\

\noindent\textbf{Cleaning and Filtering \gptfour's Responses.}
\gptfour generated responses in JSON format (example output available in supplementary materials), were turned into tabular form by extracting all the scores into individual columns along with a binary variable indicating whether a symptom was found to be explicit by \gptfour. 
The excerpts from the user-written text identified by \gptfour as explicit symptom markers were extracted using simple regular expressions. \\

\gptfive also used the same experimental design and processing steps.  

\noindent\textbf{Expert Judgments.}
Two psychology experts independently assessed the depression severity from the depression texts on a random subsample of 209 instances.
They were instructed to read the essay to identify excerpts indicative of \phq symptoms and estimate a severity score for them. 
If there were no indicative excerpts, they were asked to estimate the symptom severity based on the essay and the estimated explicit symptom scores. 
The symptoms accompanied by excerpts would pertain to explicit markers as identified by experts. \\

\subsection{Analytic Plan}
Our analytical approach draws from measurement theory and behavioral analysis to systematically understand \gptfour's and \gptfive's representation of depressive symptoms.
We used three strategies to bring out different aspects of how the models processes and integrates depressive symptoms from natural language.
We begin by assessing the convergent validity between \gptfour and \gptfive estimates, self-reported \phq scores, and expert judgments using Pearson correlations to establish the basic reliability of language models' assessments before examining its underlying mechanisms.
Following this, the rest of the analysis in the main paper focuses on \gptfour, and \gptfive's analysis can be found in the Supplementary section.
We perform latent structure analysis to model how symptoms are related within \gptfour.
Subsequently, we use Differential Language Analysis to gain a deeper understanding of how it views depression in language, and forms severity assessments.
Finally, we model the process behind the estimation of implicit symptoms' severity based on explicit symptom estimates. 

\subsubsection{Latent Structure Analysis}

\noindent\textbf{Inter-symptom Relationships.}
To understand whether \gptfour's assessment behavior reflects symptom associations consistent with established depression research, we computed Pearson correlations between all pairs of the nine \phq symptoms for both \gptfour estimates and self-reported scores.
This reveals how \gptfour conceptualizes the relationships among different depressive symptoms compared to how individuals experience and report these relationships.
The correlation matrices allow us to identify whether \gptfour has learned symptom associations that align with or deviate from established psychometric patterns in depression research.

To contrast the inter-symptom relationships between \gptfour and self-report, we computed the difference between the inter-symptoms relationship matrices. 
For statistical robustness, this difference matrix was computed over 500 trials by bootstrapped resampling to extract the 95\% confidence interval of the differences. 
CIs that included 0 were regarded statistically insignificant difference in relationship between the two symptoms $(p>.95)$ across \gptfour and self-report.\\

\noindent\textbf{Differential Item Functioning Analysis.}
We conducted Differential Item Functioning (DIF) analysis using Item Response Theory to compare how effectively each symptom discriminates between varying levels of depression severity across \gptfour estimates versus self-reported scores.
In IRT, data from ordinal items are modeled using latent variables with logistic link functions~\cite{steinberg2006using}.
For this analysis, we used the 1-parameter logistic IRT model (Rasch model)~\cite{rasch1993probabilistic} which constrains the slope parameter to 1.0, with item location parameters estimated from responses.
The item location parameter indicates where each symptom functions most effectively along the latent depression continuum.
DIF analysis examines whether items function differently across groups by comparing rank orderings of item locations.

Lower ranks indicate symptoms that discriminate better at lower depression severity levels, while higher ranks indicate symptoms most informative for distinguishing individuals with severe depression.
We computed DIF ranks for both \gptfour and self-report assessments, with rank differences revealing whether \gptfour's language-based assessment emphasizes different symptoms than traditional self-report measures across the depression severity spectrum.
To ensure statistical robustness, we estimated 95\% confidence intervals around rank differences using bootstrapped resampling over 500 trials, allowing us to identify statistically significant differences in how symptoms function across assessment methods.

\subsubsection{Explicit Symptom Identification and Linguistic Analysis}
\noindent\textbf{Language Analysis of Depression Markers.}
Beyond examining \gptfour's numerical estimates, we investigated whether the model's scoring behavior aligns with linguistic patterns known to characterize depression in human language, particularly focusing on language features that extend beyond \gptfour's explicitly identified symptom markers.
For each participant's text, we computed 1-, 2-, and 3-gram relative frequencies, categorizing tokens into two groups: those present within \gptfour's explicit spans (text segments the model cited as evidence) and those outside these spans.
To ensure robust analysis, we filtered out N-grams used by fewer than 10 participants, and removed any non alphabetic characters (including numbers or special characters).
We then computed Cohen's \textit{d}~\cite{cohen2013statistical} effect sizes comparing n-gram relative frequencies within explicit spans versus the remainder of user text.
This analysis identifies linguistic features that \gptfour systematically attributes as an explicit marker of depression.

To control for multiple comparisons, we applied Benjamini-Hochberg correction, which maintains acceptable false discovery rates when testing numerous linguistic features simultaneously.
We further refined the analysis by filtering features using Pointwise Mutual Information~\cite{church-hanks-1990-word} with a threshold of 4.0, which identifies n-grams with strong associative relationships to explicit symptom markers while removing spurious associations.
Finally, we correlated the relative frequencies of these linguistically significant features with \gptfour's total \phq scores to determine whether the model's severity estimates rely on language patterns beyond its explicitly cited evidence.

\subsubsection{Mechanism Analysis: Explicit-Implicit Symptom Interplay}
\noindent\textbf{Multivariate Regression Modeling of Symptom Inference.}
To understand \gptfour's mechanism for estimating symptoms that are not explicitly mentioned in user text, we modeled how the \gptfour's implicit symptom scores relate to its explicit symptom identifications.
For each symptom $i$ from the complete set of \phq symptoms $P$, we fitted a separate multivariate linear regression model predicting \gptfour's estimated score for symptom $i$, specifically in cases where that symptom was not explicitly identified by the model. 

The regression models used scores of all other explicitly identified symptoms as predictors, creating a comprehensive picture of how \gptfour integrates explicit information to infer unmentioned symptoms.
When a predictor symptom was not explicitly identified, we replaced its score with the dataset mean for that symptom.
$\beta$ represent the strength and direction of association between each explicit symptom and the implicit estimation of the target symptom:
\begin{equation}
    s_i^{imp} = \sum_{\substack{j \in P \\ j \neq i}} \beta_j s_j^{exp}
\end{equation}
This approach reveals \gptfour's learned patterns about symptom co-occurrence, similar to how a clinician might infer risk for sleep problems when someone explicitly reports severe mood symptoms.
To ensure statistical robustness and account for sampling variability, we repeated this entire modeling process 500 times using bootstrapped samples of the data.
Coefficients whose 90\% confidence intervals included zero were considered statistically non-significant and excluded from interpretation, providing conservative estimates of \gptfour's symptom inference mechanisms.

\section{Results}\label{sec:results}

\gptfour analyzed language samples describing depressive feelings from 955 individuals to estimate symptom-level scores based on the \phq questionnaire~\cite{kroenke2001phq}. 
First, the model identified explicitly described symptoms, citing relevant text spans from the language sample and assigning severity scores. 
For the remaining \phq symptoms that were not explicitly mentioned, \gptfour was instructed to provide rationales along with its inferred severity estimates. 
Participants independently completed the \phq questionnaire reflecting symptom severity over the previous two weeks. 
Additionally, two psychology experts manually annotated symptoms in a random subsample of 209 responses following the same guidelines given to \gptfour. 
\gptfour estimates were compared against self-reported scores on the full dataset and against expert annotations in the subsample (\autoref{tab:item_item_corr}).

\paragraph{\textbf{Convergent Validity: \gptfour estimates correlate with experts-judgments and self-report.}} 
\autoref{tab:item_item_corr} shows the average Pearson correlation ($r$) between \gptfour estimates and experts for each symptom, as well as the correlation between \gptfour and self-reported scores, both computed on the subset assessed by the experts ($N=209$).
\gptfour demonstrated high agreement with self-report ($r=0.70$) and experts-judgments (avg $r=0.81$) on total \phq ratings. 
Across all symptoms, the assessors of depression essays (i.e., \gptfour and the experts) demonstrated higher agreement with each other than either did with self-reported \phq scores. 
The absolute difference between \gptfour's agreements with self-report and experts (difference between first two columns) was less than $0.10$ for all somatic symptoms (fatigue, appetite, and psychomotor agitation/retardation), except for issues related to Sleep ($\Delta=0.15$). 
All cognitive and emotional symptoms except for anhedonia had a difference in correlation scores over 10 Pearson points. 
Highest differences were seen with suicidal ideation ($\Delta=0.32$), worthlessness/guilt ($\Delta=0.20$) and concentration difficulties ($\Delta=0.14$).  
\gptfour's agreement with experts ($r_{exp1}=0.80, r_{exp2}=0.81$) on total \phq score was very similar to the agreement between the experts ($r=0.83$).
\gptfive showed consistently lower agreement with self-report than \gptfour across all symptoms and overall \phq score.

\paragraph{\textbf{Symptom-Symptom Relationships: \gptfour and Self-report have similar inter-symptom covariates overall, except for two symptoms.}}
The symptom-symptom associations from \gptfour estimates in \autoref{fig:gpt4_sr_item_corr} (left) had positive Pearson correlations across the board ($r=0.23-0.78$). 
This was in agreement with the symptom-symptom associations from self-reported (right) scores ($r=0.38-0.78$). 
The effect sizes of the symptom-symptom relationships from \gptfour estimates were similar to self-report for all but two symptoms, namely, psychomotor agitation/retardation and suicidal ideation. 
\autoref{fig:diff_gpt4_sr_item_corr} clearly shows this difference between symptom-symptom associations from \gptfour and self-report is statistically significant (95\% CI doesn't include 0.0). 
\gptfour-estimated severities of psychomotor agitation/retardation had a stronger association with other depressive symptoms than self-report. 
In contrast,  suicidal ideation was less related to other depressive symptoms for \gptfour than in the self-report. 
This is evident from the significantly lower association of suicidal ideation with the rest of the symptoms than what was observed in self-reported scores. 

\paragraph{\textbf{Differential Item Functioning Analysis: \gptfour and Self-report have similar discriminant characteristics over latent depression scores.}}
We conducted Differential Item Functioning (DIF) analysis using Item Response Theory (IRT) to compare discriminative power between self-report and \gptfour-derived symptom scores~\cite{stark2006detecting, santelices2012relationship, belzak2020testing}. 
DIF ranks reflect how effectively each symptom distinguishes between varying levels of latent depression severity: lower ranks indicate better discrimination at lower severity, and higher ranks indicate better discrimination at higher severity. 
A positive rank difference (\gptfour minus self-report) implies that \gptfour's language-based assessment discriminates better at higher depression severity levels compared to self-report, while a negative difference indicates the opposite. 
DIF ranks for both methods, along with their 95\% confidence intervals estimated via bootstrapped resampling over 500 trials, are presented in \autoref{tab:irt}.

Overall, the discriminative characteristics of items were consistent between \gptfour and self-report assessments. Suicidal ideation, appetite, and psychomotor agitation/retardation had higher ranks (7–9), indicating these symptoms were most informative in distinguishing individuals with higher latent depression severity. 
On the other hand, anhedonia, depressed mood, and fatigue had lower ranks (1–3), signifying these symptoms effectively discriminated between individuals experiencing milder or lower levels of depression. 
All item ranks showed narrow 95\% confidence intervals, reflecting high reliability. 
Suicidal ideation (rank = 9), sleep (rank = 4), and anhedonia (rank = 3) exhibited identical ranks between \gptfour and self-report. 
Rank differences for other symptoms were minor, reinforcing that language-based severity scores from \gptfour closely align with the discriminative item characteristics observed in self-reported \phq scores.

\paragraph{\textbf{Explicit vs. Implicit Symptoms: \gptfour Estimates Explicitly Mentioned Symptoms More Precisely.}} 
Agreement between \gptfour estimates and self-report was consistently higher when symptoms were explicitly mentioned, showing an average increase of $+0.18$ Pearson correlation points over implicitly inferred symptoms --— even though explicit mentions were identified in only about one-quarter of instances (\autoref{tab:exp_imp_convergence}). 
Notably, explicit mentions significantly improved \gptfour's accuracy ($\geq0.25$ Pearson points) in estimating anhedonia, depressed mood, and suicidal ideation compared to implicit inferences. 
Language explicitly describing these three symptoms enabled more precise estimates from \gptfour, despite substantial variation in how frequently they were identified to be explicitly mentioned (range: 5.55\%–96.44\%). 
The cardinal symptoms, depressed mood and anhedonia, had particularly high rates of explicit mentions (96\% and 53\%, respectively), considerably surpassing other symptoms.

\paragraph{\textbf{Uncovering \gptfour's language of depression: \gptfour's cited evidences were reliable}}
The left bar chart of \autoref{fig:words_exp_severity_gpt4} shows that words identified by GPT-4 as explicit markers of symptoms (positive Cohen's $d$ values). 
These include emotion words (e.g., "depressed," "sad," "down"), verbs (e.g., "do"), and somatic descriptors (e.g., "tired," "sleep," "energy"). 
In contrast, words that GPT-4 rarely cited as explicit markers (negative Cohen's $d$) were mainly neutral functional words and nouns (e.g., "pandemic," "home"). 
All these highlighted words showed statistically significant associations after correcting for multiple comparisons (Benjamini-Hochberg corrected, $p<0.05$).

The bar chart on the right further shows that 10 of the 17 explicitly cited words significantly correlated with GPT-4's estimated total \phq scores ($p<0.05$). Phrases such as "\textit{have been depressed}," "\textit{sleep}," and "\textit{low}" (in "\textit{low energy}") positively correlated with higher severity scores, while phrases like "\textit{not been depressed}" and "\textit{happy}" negatively correlated, indicating lower severity. Among words not explicitly cited, only "\textit{it}" showed a small but significant correlation ($r=0.09$), suggesting GPT-4 primarily bases its severity scores on explicitly cited meaningful linguistic markers.

\paragraph{\textbf{Interplay between explicitly mentioned symptoms and inferred symptoms.}}
We examined how explicit symptoms influence implicit symptoms in \gptfour's assessments by calculating the regression coefficients ($\beta$) from a multivariate linear regression model.
This linear model predicted implicit symptom scores based on explicit symptom scores while controlling for all other implicit symptoms.
\autoref{fig:exp-imp-gpt4} depicts the effect sizes of explicit symptoms in influencing \gptfour's estimate of implicit symptoms.  
 
When depressed mood was not explicitly mentioned, it was largely inferred from explicit mentions of anhedonia ($\beta=.60$). 
Conversely, anhedonia was influenced by depressed mood ($\beta=.49$) and concentration difficulties ($\beta=.24$). 
The cardinal symptoms of depression were common explicit drivers of nearly all other symptoms with large effect sizes, underscoring their central role in the network of depressive symptoms. 
This shows that depressed mood and anhedonia are pivotal for accurately estimating the severity of other less specific symptoms of depression, such as sleep disturbances, appetite changes and fatigue, particularly when they are not directly mentioned in the text.

Furthermore, when individuals mentioned sleep disturbances, \gptfour moderately used this to estimate other somatic symptoms, such as fatigue ($\beta=.18$) and abnormal appetite ($\beta=.24$). 
However, the reverse relationship wasn't significant: sleep disturbances were not inferred from the other somatic markers. 
However, other physical symptoms didn't strongly predict sleep disturbances. Instead, sleep issues were primarily inferred from depressed mood ($\beta=.26$).

All symptoms except for fatigue that were explicitly mentioned more than 10\% of the time —-- including the cardinal symptoms and feelings of worthlessness/guilt —-- significantly helped \gptfour to estimate the severity of unmentioned symptoms.
The explicit mention of nearly all the items predicted fatigue. 
However, fatigue was a very weak indicator of other unmentioned symptoms, despite being explicitly mentioned in 1 out of 4 cases.
This reflects fatigue’s nonspecific nature, as it occurs broadly across many conditions and isn't uniquely informative for depression severity.
Three pairs of symptoms showed strong two-way relationships, each explicitly mentioned symptom significantly influencing the other’s implicit estimation and vice versa: anhedonia and depressed mood, fatigue and psychomotor agitation/retardation, and anhedonia and concentration difficulties.

\section{Discussion}\label{sec:disc}

Receiving professional mental health support is a multi-step, non-linear process that begins with screening and initial assessment. 
A significant percentage of individuals do not receive timely support due to social barriers and infrastructural limitations within the traditional mental healthcare system~\cite{coombs2021barriers, aguirre2020barriers}. 
Though effective mental health treatments exist, most individuals with mental disorders do not receive treatment, in part due to shortages of mental healthcare providers~\cite{kazdin2017addressing, konrad2009county}.
Digital technologies like \gptfour which possess unprecedented human-comparable skills and may soon be able to practically address many such challenges. 
With current consumer costs being around \$20/month, models such as \gptfour may be well-positioned to mitigate well-established inhibitors such as high costs, limited mobility, inadequate insurance coverage, and lack of support for parents~\cite{stade2024large}. 
Additionally, it can overcome infrastructural limitations, including the need for cultural adaptation and the dearth of mental health clinicians. 


While studies have focused on identifying and measuring desirable behaviors of \llm s for mental health support~\cite{chiu2024bolt}, the current study instead sought to inform how the internal structure of depression is represented within an LLM by analyzing its behavior with respect to symptoms.
Hence, we tasked \gptfour with identifying the key symptomatic markers of depression expressed within human essays.
These essays contained exhibits of causes (e.g. substance use, sedentary lifestyle) and effects (e.g. tired, exhausted) of depression through direct or synonymous mentions (e.g. ``\textit{I don't like doing things anymore}'' vs ``\textit{I don't see the point of getting out of the bed anymore}'' both suggest lack of interest).
From GPT-4's assessments on language describing depression, we learn how it views the condition to manifest in people, akin to how practicing therapists understand conditions based on the way patients describe it~\cite{frances1995dsm, sommers2012clinical}.     
We examined the feasibility of using \gptfour for depression assessment by evaluating two fundamental psychometric properties: reliability and validity. 
\gptfour demonstrated strong concurrent validity, indicating its assessments align closely with established measures of depression, such as the \phq questionnaire. 
Additionally, \gptfour showed high internal consistency, suggesting it reliably captures the interconnected nature of depressive symptoms. 
Our findings align with previous research showing that \llm s~\cite{xu-etal-2024-mental, galatzerlevy2023capability} generally correlate well with standard depression scales, and we extend these results to symptom-level assessments.
Importantly, \gptfour's cited explicit markers demonstrated both validity and reliability: the linguistic markers it identified aligned with prior findings~\cite{eichstaedt2018facebook, stade2023depression}, and were positively associated with its severity estimates, whereas uncited language from depression essays had no meaningful associations.
Such language-based assessments offer richer and more detailed insights than traditional rating scales~\cite{kjell2022natural}, which provide avenues for making hypotheses about predisposing, precipitating, and perpetuating factors of psychopathology enabling a more personalized treatment planning. 

Going beyond evaluating \gptfour's convergence with self-reports and experts, we inferred its schema of symptoms from its behavior in assessing depression.
This schema, a representation of the psychological constructs, that was derived from \gptfour's responses, was depicted as a web of associations between symptoms.   
We found \gptfour and self-report shared commonalities in their schema across the first seven \phq symptoms, while they differed for \textit{psychomotor agitation/ retardation} and \textit{suicidal thoughts}.

Specifically, \gptfour’s assessment of suicidal thoughts closely matched expert evaluations and explicit self-reports but struggled to infer it when not directly mentioned. 
This difficulty highlights the challenge of inferring severe but less frequently mentioned symptoms solely from associated emotional states like depressed mood or anhedonia.
In contrast, \gptfour consistently showed lower convergence in assessing psychomotor disturbances, suggesting that these symptoms are less observable through language compared to others. 
This finding aligns with prior research indicating that objective behavioral signals, such as speech patterns or facial expressions, are more reliable for assessing psychomotor symptoms due to potential recall biases in self-reports~\cite{Sobin1997Psychomotor}.
Notably, both suicidal ideation and psychomotor disturbances demonstrated high discriminant power with respect to latent depression severity (\autoref{tab:irt}), underscoring their clinical importance and the need to better understand how \gptfour diverges in modeling these constructs.

\begin{figure}[!ht]
    \centering
     \caption{\textbf{Effect of explicit symptom scores on implicit symptom scores as estimated by \gptfour.}
     Wider bands represent larger effect sizes ($\beta$ values, i.e., coefficients of multivariate linear regression from predicting estimated implicit symptom score using all explicit symptom scores while holding other implicit symptoms constant). 
     The 90\% confidence interval of the regression coefficients was estimated using bootstrapped resampling over 500 trials.
     Confidence intervals that included 0.0 were dropped. 
     \textit{Explicit} refers to the symptoms estimated by \gptfour when it identified explicit mention in essays and \textit{Implicit} refers to the symptom scores when they were unmentioned in essays.
     Dotted bands represent a negative association.
     }
    \includegraphics[width=\textwidth]{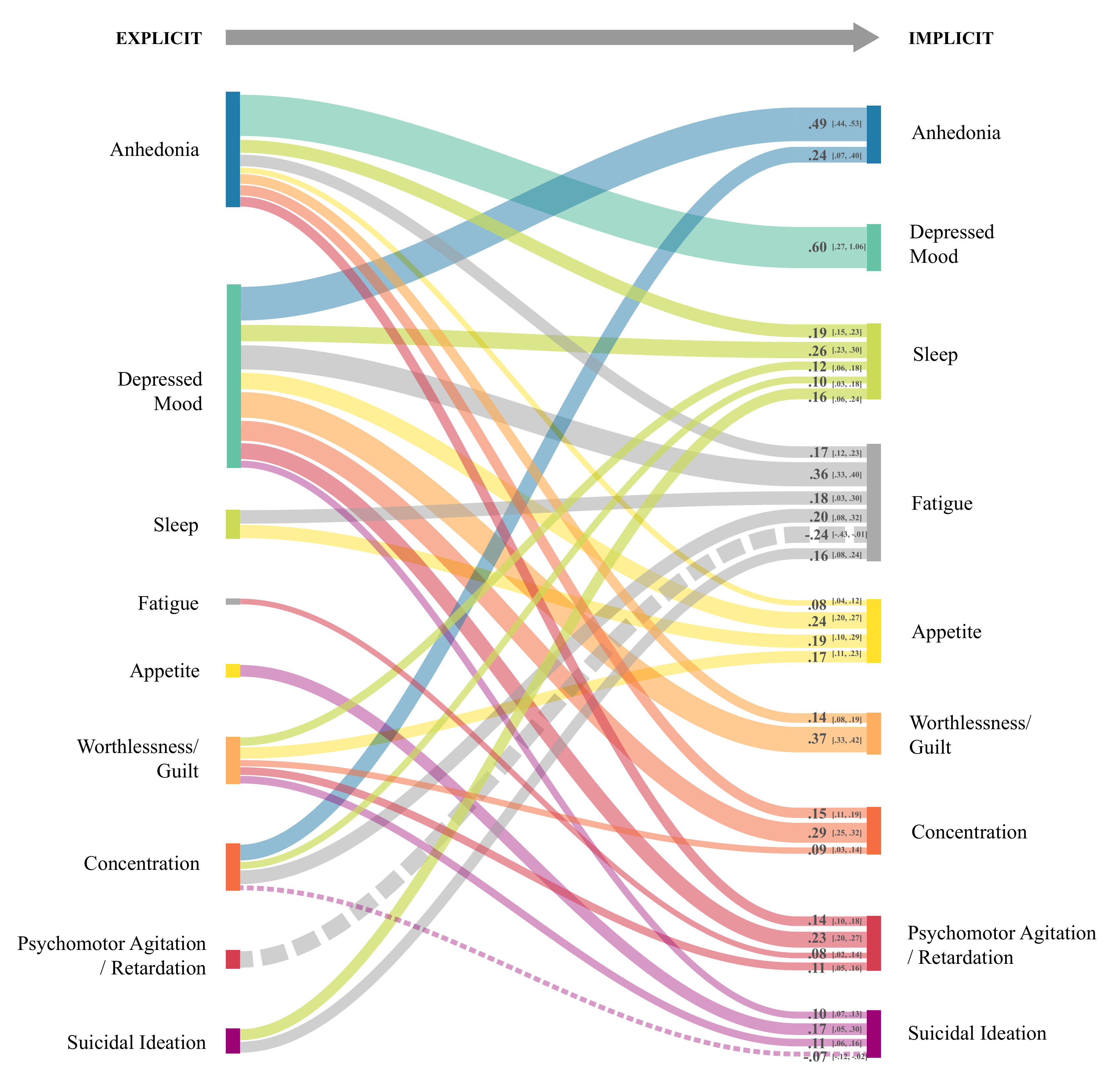}

    \label{fig:exp-imp-gpt4}
\end{figure}

\gptfour’s approach to inferring symptoms that are not explicitly mentioned aligns with prior research showing that some symptoms are more central and uniquely tied to depression, while others are broader and commonly appear across different mental health conditions~\cite{watson2009differentiating}.
Highly depression-specific symptoms such as anhedonia, depressed mood, and feelings of worthlessness/guilt~\cite{watson2009differentiating, harrison2022psychopathology} were frequently mentioned in language (\autoref{tab:exp_imp_convergence}) and strongly influenced \gptfour's inference on other symptoms (\autoref{fig:exp-imp-gpt4}).
This pattern suggests these core symptoms serve as foundational indicators in \gptfour’s schema of depression, mirroring clinical practice where these cardinal symptoms heavily guide diagnostic reasoning~\cite{fried2016good}. 
Conversely, nonspecific symptoms, such as sleep disturbances and appetite changes -- common to both depression and anxiety disorders -- could be reliably inferred by \gptfour from the presence of almost any explicitly mentioned symptom.

Another important consideration in explaining LLMs is the instructions it was given. 
As the information in instructions can affect the model's response~\cite{ceballos-arroyo-etal-2024-open}, we instructed \gptfour to estimate symptoms of depression before the overall depression severity. 
This provided a means to understand the cohesiveness of the structure of depression, its faithfulness to what it identifies as explicit markers, and varying difficulty levels in estimating the symptoms.
This mirrors studies of LLMs in other domains that have demonstrated improved effectiveness when allowed to break a task into compositional steps~\cite{wei-etal-2022-chain} or when instructed to carry out specific modules that culminate into a task~\cite{khot2023decomposed}. 

There has been increased advocacy for transparency and explainability~\cite{WHO2021_AI_Health_Ethics} to improve public faith and understand the risks it comes with, especially in areas such as mental health support~\cite{lee2021artificial}. 
Using measurement theory to arrive at explanatory schemas can inform how LLMs function, where they are bound to fail, and how to effectively situate them in the mental health support to maximize overall benefits.  
Our evaluation framework is intended to help clinicians and decision-makers gain insights about model capabilities and pitfalls to determine the capacity at which contemporary models can be used in the care pipeline and further improve future systems to yield stronger outcomes. 

Our work had several limitations.
Foremost, the scope of our results is limited to explaining \gptfour's behaviors when depression is conceptualized through the lens of the \phq.
While the \phq is a widely used and validated tool, it reflects a categorical, DSM-based view of depression that may not fully capture the complexity of psychological distress. 
Future work can expand this approach to data-driven, dimensional models such as HiTOP~\cite{kotov2017hierarchical}, which may better capture the heterogeneity and complexity of psychopathology across individuals.
Although other modalities beyond language (e.g., speech, behavior) can enrich assessment, we focus on language due to its information-rich nature and the ability of LLMs to leverage this medium for making  accurate assessments~\cite{kjell2022natural, varadarajan-etal-2024-alba}.
Additionally, our study was limited to a cross-sectional analysis of a relatively small, predominantly UK-based sample. 
Broader validation across diverse populations and contexts is necessary.
Accurate mental health assessments would require taking longitudinal symptom trajectories into account ~\cite{galatzer-levy-etal-2023-machine} to overcome the limitations of cross-sectional evaluations, which fails to capture the dynamic nature of symptom progression and individual variability over time. 
Although transformer-based models have been effective in encoding longitudinal representations of behavioral~\cite{xu2022globem} and health markers~\cite{theodorou2023synthesize}, it is vital to understand how effective are LLMs in disentangling both these trajectories over time to assess the mental health.  

\section{Conclusion} \label{sec:conclusion}
This study uncovered the underlying symptom schema within \gptfour that explains how its arrived at scores having high convergence with experts and standard self-report tests.
While \gptfour’s schema closely matched self-report for major symptoms, it notably downplayed the link with suicidal ideation and overemphasized psychomotor agitation/retardation. 
This divergence was somewhat mitigated when suicidal ideation was explicitly mentioned, suggesting it is more conservative when inferring symptoms of suicidality without explicit evidence. 
On the other hand, its overemphasis on psychomotor symptoms seemed to blanket bias, happening whether it identified evidence or not. 
With awareness of these discrepancies, the results suggest \gptfour has the potential to assist clinicians, such as in screening to optimize time for those in urgent need. 
More widely, this work demonstrates how a machine behavior approach yields explanations of large language models' processing of psychological concepts.

\subsection*{Ethics Approval and Consent to participate}
This study was deemed exempt from IRB approval. All study participants' data was anonymized and they consented to share the data for research. 

\subsection*{Declarations}
\textbf{Funding. }
AVG and RLB were supported in part by NIH grant R01-AA028032, VV was supported in part by a DARPA grant \#W911NF-20-1-0306, and HAS was supported in part by both the grants. 
YKL was supported in part by the Air Force Research Laboratory (AFRL), DARPA, for the KAIROS program under agreement number FA8750-19-2-1003 and in part by the NSF under the award IIS \#2007290. 
JCE was supported by Stanford's Institute for Human-Centered AI.
LF was supported by the German Federal Ministry of Education and Research (BMBF) as a part of the AI Research Group program under the reference 01-S20060, by the state of North Rhine-Westphalia as part of the Lamarr Institute for Machine Learning and Artificial Intelligence research on LLMs, and by the Bonn-Aachen International Center for Information Technology (b-it) supporting the visiting research stay at Stony Brook University. ONEK, KK and VCE were supported by FORTE (STY-2022/0007; 2022-01022).

The conclusions contained herein are those of the authors and should not be interpreted as necessarily representing the official policies, either expressed or implied, of DARPA, NIH, any other government organization, or the U.S./ German Government.

\noindent\textbf{CoI/Competing Interest. } 
ONEK and KK co-founded a start-up that uses computational language assessments to diagnose mental health problems.


\subsection*{Preprints}
A preprint of this article is available online: \href{https://arxiv.org/abs/2411.13800}{arXiv.2411.13800}

\subsection*{Data availability}
The data for this project will be made available through Github.
All data including the original study data, \gptfour responses, cleaned version, as well as the processed features will be open-sourced for replication. 
All programs for this work were coded in Python and will be open-sourced with the data in the same Github repository. 
This includes programs that accessed \gptfour to generate responses using OpenAI's API, cleaning the data, and our analysis.    

\subsection*{Acknowledgments}
The authors extend their gratitude to Daniel Klein, Roman Kotov, Nicholas Eaton, Dina Vivian, Whitney Ringwald, August Nilsson, and Siddharth Manglik for providing feedback on this work.

\medskip

\bibliography{references} 

\section{Tables \& Figures}

\begin{table}[!ht]
    \centering
    \caption{\textbf{Overall and Symptom-level Convergent Validity.} Overall PHQ9 scores as well as item-item Pearson correlation ($r$) of \gptfour with experts' judgments (averaged correlations) (left) and the self-reported scores (second-left), \gptfive with self-reported scores (second-right) and averaged correlations between experts and self-report (right) ($N=209$). Statistically significant differences from the self-report-\gptfour agreement are indicated by $\dagger$ $(p<.05)$ and $\ddagger$ $(p<.01)$, computed using bootstrapped resampling across individuals over 500 trials. }
    \resizebox{\columnwidth}{!}{%
    \begin{tabular}{lcccc}
    \toprule    
     & \textbf{Experts - \gptfour} & \textbf{Self-report - \gptfour} & \textbf{Self-report - \gptfive} & \textit{Experts - Self-report} \\
    \textbf{Symptom} & \textbf{$r$} & \textbf{$r$} & $r$ \\
    \midrule
    PHQ 9 Total & .81\hspace{.1cm} & .70 & .68 & \textit{.68} \\
    \midrule
    Anhedonia & .65\hspace{.1cm} &  .59 & .57 & \textit{.56} \\
    Depressed & .81$^\dagger$ & .69 & .68 & \textit{.66} \\
    Sleep & .69$^\dagger$ & .54 & .46 & \textit{.46} \\
    Fatigue & .68$^\dagger$ & .59 & .47 & \textit{.53}\\
    Appetite & .49\hspace{.1cm} & .43 & .33 & \textit{.43}\\
    Worthlessness/Guilt & .70$^\dagger$ & .50 & .40 & \textit{.53}\\
    Concentration & .62$^\dagger$ & .48 & .33 & \textit{.45}\\
    Psychomotor  & .31$^\dagger$ & .23 & .14 & \textit{.14}\\
    Suicidal Thoughts & .63$^\ddagger$ & .31 & .25 & \textit{.30}\\
    \bottomrule
    \end{tabular}
    }
    \label{tab:item_item_corr}
\end{table}

\definecolor{rank1}{HTML}{e06666}
\definecolor{rank2}{HTML}{e67f73}
\definecolor{rank3}{HTML}{ec9881}
\definecolor{rank4}{HTML}{f2b18e}
\definecolor{rank5}{HTML}{f9cb9c}
\definecolor{rank6}{HTML}{fad4a8}
\definecolor{rank7}{HTML}{fcdeb4}
\definecolor{rank8}{HTML}{fde8c0}
\definecolor{rank9}{HTML}{fff2cc}

\begin{figure}[!t]
\centering
\caption{\textbf{Similarities and differences between \gptfour's and self-report's schema of depressive symptoms.}}

\subfloat[\textbf{Pearson correlation between symptoms estimated by \gptfour (left) and self-reported by participants (right).} The average correlations of each symptom with rest are presented as column vectors for both \gptfour and self-report. N=955. ]{
    \includegraphics[width=\linewidth]{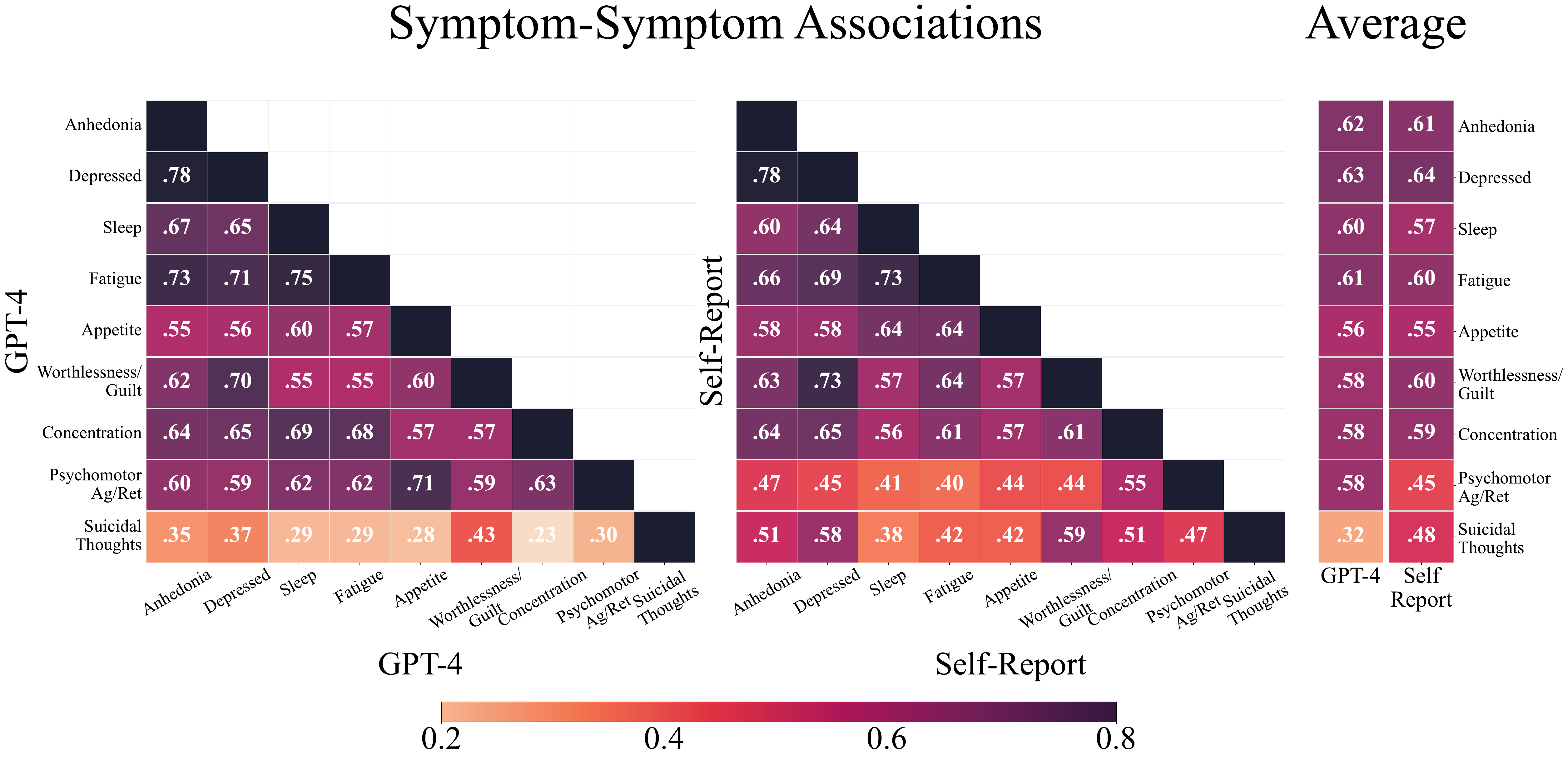}
    \label{fig:gpt4_sr_item_corr}
}
\par  
\subfloat[\textbf{Difference between \gptfour's and self-reports' Symptom-Symptom correlations.} 95\% confidence interval of the difference in correlations from bootstrapped resampling over 500 trials is enclosed within brackets. \gptfour's symptom-symptom relationship is largely similar to self-report, except for psychomotor agitation/retardation and suicidal ideation.]{
    \includegraphics[width=0.50\linewidth]{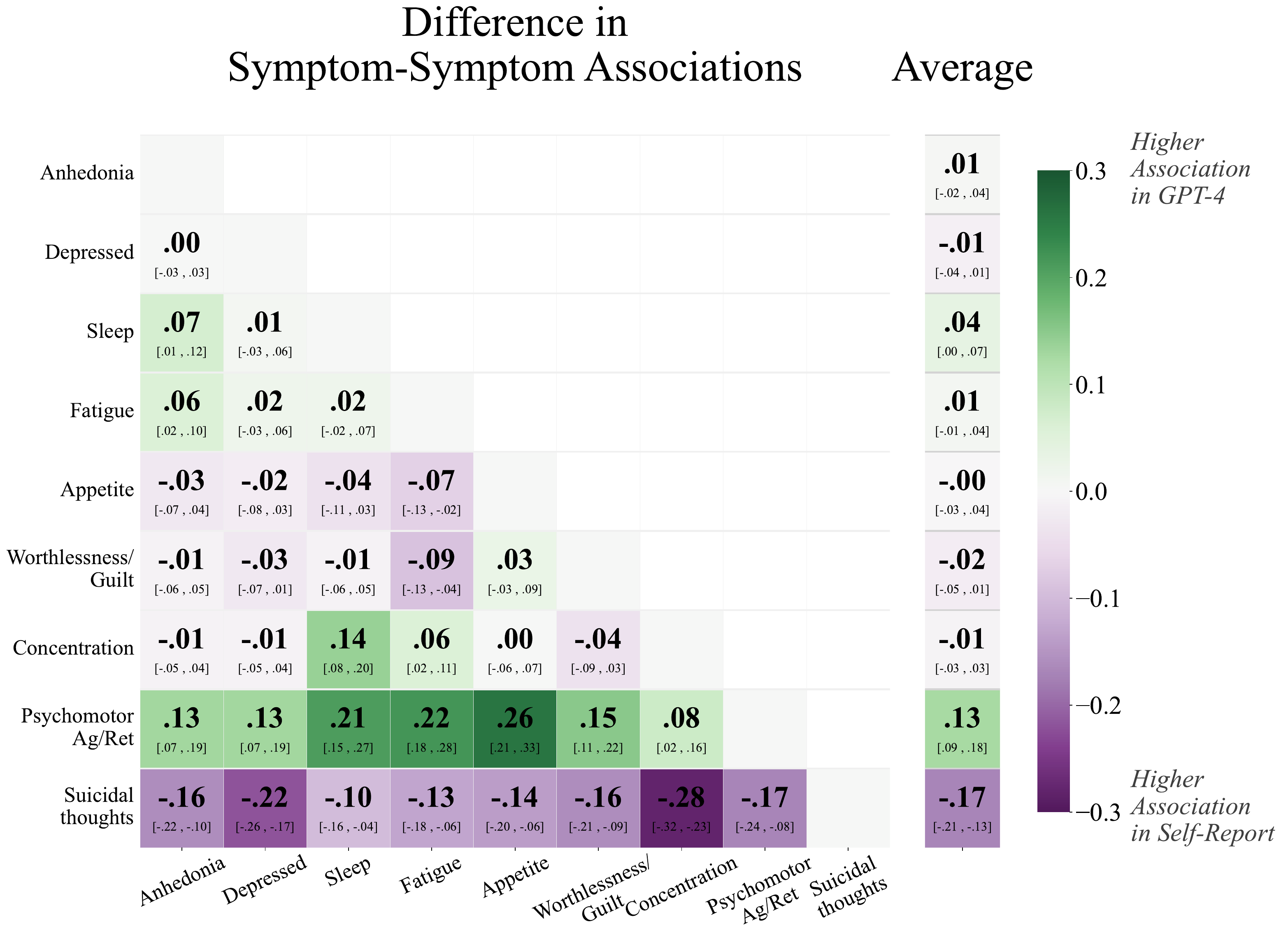}
    \label{fig:diff_gpt4_sr_item_corr}
}
\hspace{2pt}
\subfloat[\textbf{Symptom's item location ranking according to \textit{IRT}} over \gptfour and self-report scores for zero and non-zero decision boundary. 
    A lower rank value corresponds to the symptom being best at distinguishing lower values of depression, while a higher rank indicates being better at distinguishing higher values of depression. 
    Ranks are accompanied by a 95\% confidence interval estimated using bootstrapped resampling across individuals (N=955, num. trials=500).]{
    \begin{minipage}{0.45\linewidth}
    \fontsize{10pt}{9pt}\selectfont
    \begin{tabular}{l|P{1cm}P{1cm}P{1cm}}
    \hline
    \textbf{PHQ9 Item}   & \multicolumn{3}{c}{\textbf{Item Location Rank}}      \\
           & \textbf{\gptfour}    & \textbf{Self-Report}  & \textbf{Item location diff}  \\
    \hline
    Anhedonia & \cellcolor{rank3} 3 \tiny{[1, 3]} & \cellcolor{rank3} 3 \tiny{[3, 3]} & 0 \\ 
    Depressed Mood & \cellcolor{rank1} 1 \tiny{[1, 2]} & \cellcolor{rank2} 2 \tiny{[1, 2]} & -1 \\ 
    Sleep & \cellcolor{rank4} 4 \tiny{[4, 5]} & \cellcolor{rank4} 4 \tiny{[4, 5]} & 0 \\ 
    Fatigue & \cellcolor{rank2} 2 \tiny{[1, 3]} & \cellcolor{rank1} 1 \tiny{[1, 2]} & 1 \\ 
    Appetite & \cellcolor{rank8} 8 \tiny{[7, 8]} & \cellcolor{rank7} 7 \tiny{[7, 7]} & 1 \\ 
    Worthlessness / Guilt & \cellcolor{rank6} 6 \tiny{[6, 6]} & \cellcolor{rank5} 5 \tiny{[4, 6]} & 1 \\ 
    Difficulty concentrating & \cellcolor{rank5} 5\tiny{[4, 5]} & \cellcolor{rank6} 6\tiny{[4, 6]} & -1 \\ 
    Psychomotor & \cellcolor{rank7} 7\tiny{[7, 8]} & \cellcolor{rank8} 8\tiny{[8, 8]} & -1 \\ 
    Suicidal Ideation & \cellcolor{rank9} 9\tiny{[9, 9]} & \cellcolor{rank9} 9\tiny{[9, 9]} & 0 \\ 
    \hline
    \end{tabular}
    \label{tab:irt}
    \end{minipage}
}

\end{figure}

\definecolor{barcolor}{HTML}{fdae61}
\definecolor{cbarcolor}{HTML}{9e9ac8}

\begin{figure}[!ht]
\centering
\caption{\textbf{Explicit Symptom mentions Improves \gptfour's Accuracy and Informs Severity Judgments.}}

\subfloat[
    \textbf{Contrast of explicit \& implicit symptom mentions and convergence.} Pearson Correlation ($r$) between \gptfour and self-reported scores for all symptoms under conditions where symptoms were identified to be explicit and implicit (N=955) by \gptfour. 
    The bar chart in \% Explicit column represents the rate of explicit mentions according to \gptfour. 
    Higher correlations were observed consistently when symptoms were identified to be explicitly mentioned.]{
    \begin{minipage}{\textwidth}
    \centering
    \fontsize{9pt}{4pt}\selectfont 
\begin{tabular}{m{3.5cm} >{\centering\arraybackslash}m{1.5cm} >{\centering\arraybackslash}m{5cm} >{\centering\arraybackslash}m{1.5cm}}
        \toprule
        \textbf{Symptom} & \textbf{Explicit $r$} & \textbf{\% Explicit vs. \% Implicit}  & \textbf{Implicit $r$} \\
        \midrule
        \vspace*{3pt}Anhedonia & \vspace*{3pt}.71 &  
        \begin{tikzpicture}[baseline=(current bounding box.center)]
        \begin{axis}[
            width=5cm, height=1.85cm, 
            axis lines=none, 
            xbar stacked,    
            bar width=0.4cm,   
            xmin=0, xmax=100, 
            ymin=0, ymax=1,   
            xtick=\empty,
            ytick=\empty,    
            clip=false, 
            ]
            \addplot [fill=barcolor] coordinates {(53.72,0)}; 
            \addplot [fill=cbarcolor] coordinates {(46.28,0)}; 
            \node[anchor=east, inner sep=0pt] at (axis cs:-2,0) {\makebox[15mm][r]{53.72\%}};
        \end{axis}
        \end{tikzpicture} & \vspace*{3pt}.43  \\
        \vspace*{3pt}Depressed Mood & \vspace*{3pt}.70 &  
        \begin{tikzpicture}[baseline=(current bounding box.center)]
        \begin{axis}[
            width=5cm, height=1.85cm,
            axis lines=none,
            xbar stacked,
            bar width=0.4cm,
            xmin=0, xmax=100,
            ymin=0, ymax=1,
            xtick=\empty,
            ytick=\empty,
            clip=false,
            ]
            \addplot [fill=barcolor] coordinates {(96.44,0)}; 
            \addplot [fill=cbarcolor] coordinates {(3.56,0)}; 
            \node[anchor=east, inner sep=0pt] at (axis cs:-2,0) {\makebox[15mm][r]{96.44\%}};
        \end{axis}
        \end{tikzpicture} & \vspace*{3pt}.44 \\
        \vspace*{3pt}Sleep & \vspace*{3pt}.52 & 
        \begin{tikzpicture}[baseline=(current bounding box.center)]
        \begin{axis}[
            width=5cm, height=1.85cm,
            axis lines=none,
            xbar stacked,
            bar width=0.4cm,
            xmin=0, xmax=100,
            ymin=0, ymax=1,
            xtick=\empty,
            ytick=\empty,
            clip=false,
            ]
            \addplot [fill=barcolor] coordinates {(11.31,0)}; 
            \addplot [fill=cbarcolor] coordinates {(88.69,0)}; 
            \node[anchor=east, inner sep=0pt] at (axis cs:-2,0) {\makebox[15mm][r]{11.31\%}};
        \end{axis}
        \end{tikzpicture} & \vspace*{3pt}.43 \\
        \vspace*{3pt}Fatigue & \vspace*{3pt}.58 & 
        \begin{tikzpicture}[baseline=(current bounding box.center)]
        \begin{axis}[
            width=5cm, height=1.85cm,
            axis lines=none,
            xbar stacked,
            bar width=0.4cm,
            xmin=0, xmax=100,
            ymin=0, ymax=1,
            xtick=\empty,
            ytick=\empty,
            clip=false,
            ]
            \addplot [fill=barcolor] coordinates {(25.13,0)}; 
            \addplot [fill=cbarcolor] coordinates {(74.87,0)}; 
            \node[anchor=east, inner sep=0pt] at (axis cs:-2,0) {\makebox[15mm][r]{25.13\%}};
        \end{axis}
        \end{tikzpicture} & \vspace*{3pt}.51  \\
        \vspace*{3pt}Appetite & \vspace*{3pt}.47 & 
        \begin{tikzpicture}[baseline=(current bounding box.center)]
        \begin{axis}[
            width=5cm, height=1.85cm,
            axis lines=none,
            xbar stacked,
            bar width=0.4cm,
            xmin=0, xmax=100,
            ymin=0, ymax=1,
            xtick=\empty,
            ytick=\empty,
            clip=false,
            ]
            \addplot [fill=barcolor] coordinates {(5.03,0)}; 
            \addplot [fill=cbarcolor] coordinates {(94.97,0)}; 
            \node[anchor=east, inner sep=0pt] at (axis cs:-2,0) {\makebox[15mm][r]{5.03\%}};
        \end{axis}
        \end{tikzpicture} & \vspace*{3pt}.33 \\
        \vspace*{3pt}Worthlessness / Guilt & \vspace*{3pt}.57 & 
        \begin{tikzpicture}[baseline=(current bounding box.center)]
        \begin{axis}[
            width=5cm, height=1.85cm,
            axis lines=none,
            xbar stacked,
            bar width=0.4cm,
            xmin=0, xmax=100,
            ymin=0, ymax=1,
            xtick=\empty,
            ytick=\empty,
            clip=false,
            ]
            \addplot [fill=barcolor] coordinates {(24.19,0)}; 
            \addplot [fill=cbarcolor] coordinates {(75.81,0)}; 
            \node[anchor=east, inner sep=0pt] at (axis cs:-2,0) {\makebox[15mm][r]{24.19\%}};
        \end{axis}
        \end{tikzpicture} & \vspace*{3pt}.45 \\
        \vspace*{3pt}Concentration & \vspace*{3pt}.50 & 
        \begin{tikzpicture}[baseline=(current bounding box.center)]
        \begin{axis}[
            width=5cm, height=1.85cm,
            axis lines=none,
            xbar stacked,
            bar width=0.4cm,
            xmin=0, xmax=100,
            ymin=0, ymax=1,
            xtick=\empty,
            ytick=\empty,
            clip=false,
            ]
            \addplot [fill=barcolor] coordinates {(9.42,0)}; 
            \addplot [fill=cbarcolor] coordinates {(90.58,0)}; 
            \node[anchor=east, inner sep=0pt] at (axis cs:-2,0) {\makebox[15mm][r]{9.42\%}};
        \end{axis}
        \end{tikzpicture} & \vspace*{3pt}.45  \\
        \vspace*{3pt}Psychomotor & \vspace*{3pt}.33 & 
        \begin{tikzpicture}[baseline=(current bounding box.center)]
        \begin{axis}[
            width=5cm, height=1.85cm,
            axis lines=none,
            xbar stacked,
            bar width=0.4cm,
            xmin=0, xmax=100,
            ymin=0, ymax=1,
            xtick=\empty,
            ytick=\empty,
            clip=false,
            ]
            \addplot [fill=barcolor] coordinates {(3.35,0)}; 
            \addplot [fill=cbarcolor] coordinates {(96.65,0)}; 
            \node[anchor=east, inner sep=0pt] at (axis cs:-2,0) {\makebox[15mm][r]{3.35\%}};
        \end{axis}
        \end{tikzpicture} & \vspace*{3pt}.24  \\ 
        \vspace*{3pt}Suicidal Ideation & \vspace*{3pt}.75 & 
        \begin{tikzpicture}[baseline=(current bounding box.center)]
        \begin{axis}[
            width=5cm, height=1.6cm,
            axis lines=none,
            xbar stacked,
            bar width=0.4cm,
            xmin=0, xmax=100,
            ymin=0, ymax=1,
            xtick=\empty,
            ytick=\empty,
            clip=false,
            ]
            \addplot [fill=barcolor] coordinates {(5.55,0)}; 
            \addplot [fill=cbarcolor] coordinates {(94.45,0)}; 
            \node[anchor=east, inner sep=0pt] at (axis cs:-2,0) {\makebox[15mm][r]{5.55\%}};
        \end{axis}
        \end{tikzpicture} & \vspace*{3pt}.21 \\
        \bottomrule
        \vspace*{3pt}\textbf{Mean} & \vspace*{3pt}.57 & 
        \begin{tikzpicture}[baseline=(current bounding box.center)]
        \begin{axis}[
            width=5cm, height=1.6cm,
            axis lines=none,
            xbar stacked,
            bar width=0.4cm,
            xmin=0, xmax=100,
            ymin=0, ymax=1,
            xtick=\empty,
            ytick=\empty,
            clip=false,
            ]
            \addplot [fill=barcolor] coordinates {(26.02,0)}; 
            \addplot [fill=cbarcolor] coordinates {(73.98,0)}; 
            \node[anchor=east, inner sep=0pt] at (axis cs:-2,0) {\makebox[15mm][r]{26.02\%}};
        \end{axis}
        \end{tikzpicture} & \vspace*{3pt}.39 \\
        \bottomrule
    \end{tabular}
    \label{tab:exp_imp_convergence}
    \end{minipage}
}
\par  
\subfloat[\textbf{\gptfour's Language of Depression: } 
    Words from the human-written essays associated with explicit symptomatic markers (left) and higher severity score (right) estimated by \gptfour. 
    Positive \textit{Cohen's d} values signify \gptfour-identified these words as explicit markers $(p<0.05)$. 
    Higher Pearson $r$ maps to higher depression severity and words with significant correlations $(p<0.05)$ are boldfaced. Both tests were multi test corrected using Benjamini-Hochberg procedure.]{
    \begin{minipage}{\textwidth}
    \centering
    \includegraphics[width=0.75\textwidth]{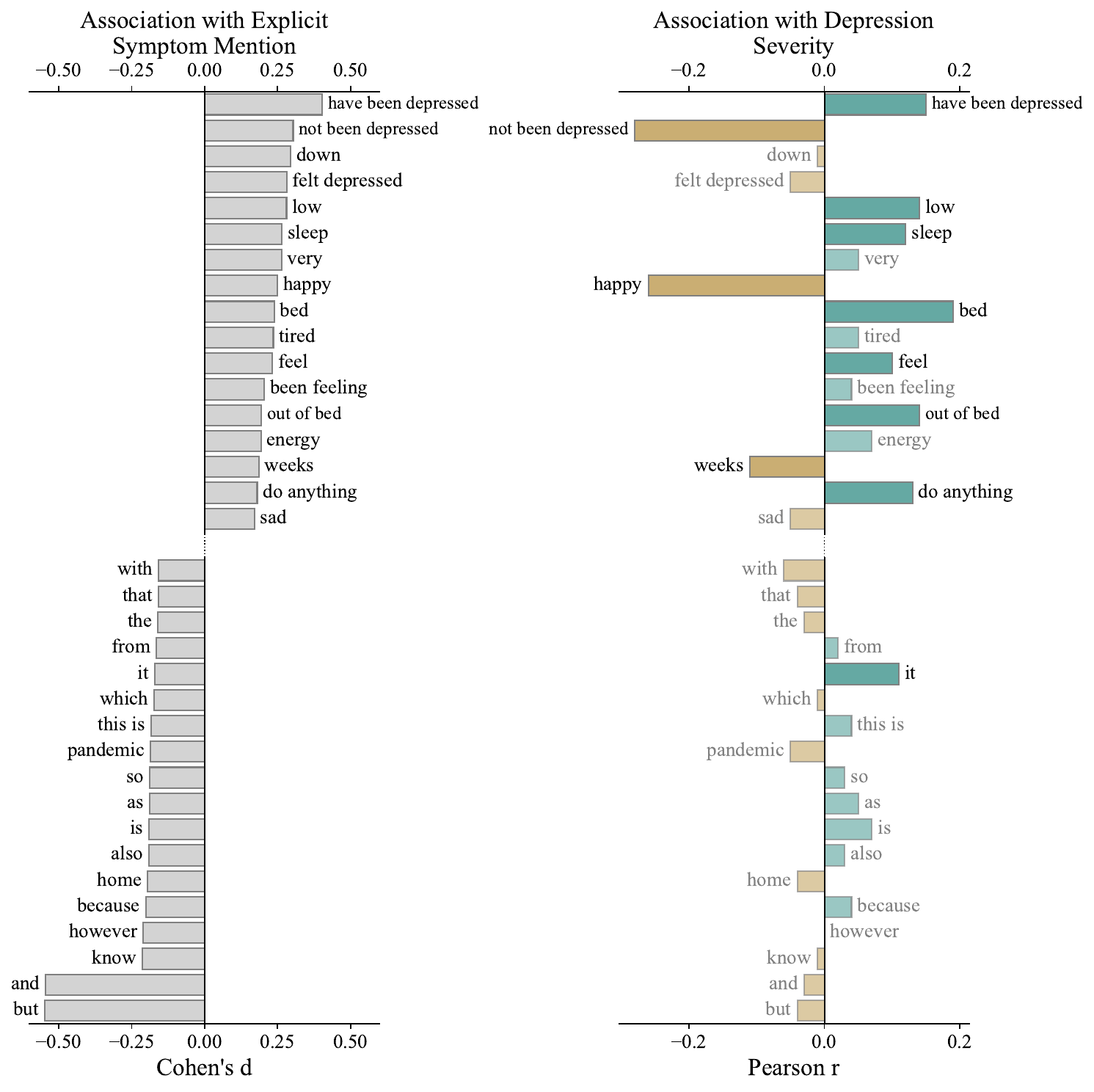}
    \label{fig:words_exp_severity_gpt4}
    \end{minipage}
}
\end{figure}

\FloatBarrier

\section*{Supplementary}

The distribution of severity scores for self-report and \gptfour's assessments over 955 samples from Gu et al., 2024~\cite{gu2024natural} is shown in ~\autoref{tab:combined_score_dist}. 
While, \gptfour's estimation rate of symptom absence (score=0) is very similar to that of self-report, its estimates were rarely 3 (most severe) and was most commonly 1 for majority of the symptoms, showing a different distribution at higher values.
This shows that \phq displays an under-estimation bias of severity scores.

\begin{table*}[!th]
\centering
\caption{Distribution of self-report and \gptfour estimated scores for each item on \phq (N=955). 
\gptfour's estimation rate of symptom absence (score=0) is very similar to that of self-report, but
its estimates were rarely 3 (most severe) and was most commonly 1 for majority of the symptoms.
}
\begin{tabular}{l|P{0.82cm}P{0.82cm}|P{0.82cm}P{0.82cm}|P{0.82cm}P{0.82cm}|P{0.82cm}P{0.82cm}}
\toprule
\textbf{Item} & \multicolumn{2}{c|}{\textbf{\% Scoring 0}} & \multicolumn{2}{c|}{\textbf{\% Scoring 1}} & \multicolumn{2}{c|}{\textbf{\% Scoring 2}} & \multicolumn{2}{c}{\textbf{\% Scoring 3}} \\
              & \textbf{Self Report} & \textbf{GPT-4} & \textbf{Self Report} & \textbf{GPT-4} & \textbf{Self Report} & \textbf{GPT-4} & \textbf{Self Report} & \textbf{GPT-4} \\
\midrule
Anhedonia                         & 24.3 & 28.5 & 32.5 & 43.4 & 24.4 & 26.5 & 18.8 & 1.7 \\
Depressed Mood                    & 23.8 & 25.0 & 32.1 & 24.6 & 22.0 & 44.0 & 22.2 & 6.5 \\
Sleep                             & 25.1 & 35.2 & 23.8 & 54.4 & 21.1 & 10.1 & 30.0 & 0.4 \\
Fatigue                           & 15.4 & 29.3 & 27.2 & 50.9 & 22.7 & 19.4 & 34.6 & 0.4 \\
Appetite                          & 30.6 & 57.2 & 25.9 & 40.0 & 19.1 & 2.7  & 24.4 & 0.1 \\
Worthlessness / Guilt            & 29.4 & 43.3 & 25.4 & 37.5 & 23.0 & 18.3 & 22.2 & 0.9 \\
Difficulty concentrating          & 28.0 & 36.0 & 30.9 & 58.3 & 24.1 & 5.7  & 16.9 & 0.0 \\
Psychomotor Agi./ Ret.            & 55.5 & 53.3 & 23.2 & 45.3 & 15.0 & 1.4  & 6.4  & 0.0 \\
Suicidal ideation                 & 60.3 & 87.9 & 21.4 & 8.8  & 10.0  & 2.3  & 8.3  & 1.1 \\
\bottomrule
\end{tabular}
\label{tab:combined_score_dist}
\end{table*}

\definecolor{Lavender}{HTML}{E6E6FA}
\definecolor{Periwinkle}{HTML}{CCCCFF}
\definecolor{SeaGreen}{RGB}{46, 139, 87}      
\definecolor{SkyBlue}{RGB}{135, 206, 235}     
\definecolor{Orange}{HTML}{fdae61}
\definecolor{Brown}{HTML}{dfc27d}

\begin{figure*}[!bh]
\centering
\small
\caption{\textbf{Prompt given to the study Participants to express their feeling of depression using language.}  
}

\begin{tcolorbox}[
    width=\textwidth,
    colback=white,
    colframe=black,
    arc=4mm,
    boxrule=0.5pt,
    left=2mm,
    right=2mm,
    top=2mm,
    bottom=2mm,
    fonttitle=\bfseries,
    ]

\begin{tcolorbox}[
    colback=Brown!8,
    boxrule=0pt,
    colframe=white,
    left=0pt,
    right=0pt,
    top=0pt,
    bottom=0pt,
    ]

\small

Over the last 2 weeks, have you been depressed or not? 
Please answer the question by typing at least a paragraph below that indicates whether you have been depressed or not. Try to weigh the strength and the number of aspects that describe if you have been depressed or not so that they reflect your overall personal state of depression. For example, if you have been depressed, then write more about aspects describing this, and if you have not been depressed, then write more about aspects describing that. 
Write about those aspects that are most important and meaningful to you. 
Write at least one paragraph in the box.

\end{tcolorbox}
\end{tcolorbox}
\label{fig:depression_prompt_participant}
\end{figure*}

\begin{figure}[!th]
\centering
\small
\caption{\textbf{Instruction to GPT-4 with an example input text.} The instruction given to \gptfour\ for depression assessment comprised three parts: a brief \hlc[SeaGreen!10]{task description}, detailed \hlc[Periwinkle!30]{step-by-step instructions}, and the \hlc[SkyBlue!20]{formatting structure}. 
\gptfour was prompted with individual depression essays along with this instruction.}
\begin{tcolorbox}[
    width=1\textwidth,
    colback=white,
    colframe=black,
    arc=2mm,
    boxrule=0.5pt,
    left=1mm,
    right=1mm,
    top=1mm,
    bottom=1mm,
    fonttitle=\bfseries,
    ]

\begin{tcolorbox}[
    colback=SeaGreen!8,
    boxrule=0pt,
    colframe=white,
    left=0pt,
    right=0pt,
    top=0pt,
    bottom=0pt,
    ]

\fontsize{9pt}{4pt}\selectfont Your task is to conduct an in-depth analysis of a provided written text, with the goal of mirroring the psychological state of the author to accurately fill out the Patient Health Questionnaire (PHQ9). The PHQ9 consists of 9 items that are designed to identify symptoms of depression:

\begin{itemize}
    \item \textbf{Anhedonia}: Little interest or pleasure in doing things
    \item \textbf{Depressed Mood}: Feeling down, depressed, or hopeless
    \item \textbf{Insomnia or Hypersomnia}: Trouble falling or staying asleep, or sleeping too much
    \item \textbf{Fatigue}: Feeling tired or having little energy
    \item \textbf{Poor Appetite or Overeating}
    \item \textbf{Worthlessness or Guilt}: Feeling bad about yourself or that you are a failure or have let yourself or your family down
    \item \textbf{Difficulty Concentrating}: Trouble concentrating on things
    \item \textbf{Psychomotor Agitation or Retardation}: Moving or speaking so slowly that other people could have noticed, or being so fidgety or restless that you have been moving around a lot more than usual
    \item \textbf{Suicidal Ideation}: Thoughts that you would be better off dead, or of hurting yourself
\end{itemize}
\end{tcolorbox}

\begin{tcolorbox}[
    colback=Periwinkle!20,
    boxrule=0pt,
    colframe=white,
    left=0pt,
    right=0pt,
    top=0pt,
    bottom=0pt,
    ]

\fontsize{9pt}{4pt}\selectfont The text should be scrutinized carefully and the following steps should be strictly followed:\\
\textbf{STEP 1}: Pinpoint and record only the PHQ9 items that are clearly stated in the text, assigning each a severity score between 0--3. At this stage, refrain from evaluating or recording items that are not explicitly mentioned. Enclose the text spans indicative of respective symptoms inside single quotes. If multiple spans are present, separate them using comma.

\textbf{STEP 2}: For the remaining PHQ9 items not explicitly mentioned in the text, assign an estimated score. This estimation should be based on what you would expect it to be while factoring the severity score of the symptoms already identified.

\textbf{STEP 3}: Combine the scores from the previous steps and present the total PHQ9 score as ``Combined Score:''. Categorize this score into only one of the following: None, Mild, Moderate, Moderately Severe, and Severe, and present it as ``Category:''. This will signify the completion of the task.

\end{tcolorbox}

\begin{tcolorbox}[
    colback=SkyBlue!10,
    boxrule=0pt,
    colframe=white,
    left=0pt,
    right=0pt,
    top=0pt,
    bottom=0pt,
    ]

\fontsize{9pt}{4pt}\selectfont To ensure clarity and easy readability, format your output into a nested JSON. The first level should contain the step number as key (`STEP \#'), and the value should be a JSON containing the items as keys and a list containing reason(s) and the estimated severity score as values. The final step's JSON should contain `Final' as the key and a list containing the summed score and the evaluated category as the value. Here's the abstracted format of the JSON, with elements inside the angle brackets being placeholders for the actual values:

\begin{verbatim}
{
    "STEP 1": {
        "<Item a>": ["<Reason>", "<Severity Score a>"],
        "<Item b>": ["<Reason>", "<Severity Score b>"],
        ...
    },
    "STEP 2": {
        "<Item x>": ["<Reason>", "<Severity Score x>"],
        "<Item y>": ["<Reason>", "<Severity Score y>"],
        ...
    },
    "STEP 3": {
        "Final": ["Combined Score: <score>", "Category: <category>"]
    }
}
\end{verbatim}

Keep in mind, a score of 0 indicates that the symptom was not experienced, while a score of 3 signifies a high severity level of the specific symptom.

\end{tcolorbox}

\fontsize{9pt}{4pt}\selectfont \textit{\textbf{Text}}: \textit{``I feel overwhelmed because I am bearing the weight of all the domestic and financial responsibilities. I should be getting more help than I am and it makes me feel resentful and annoyed with myself for allowing it to continue. I am avoiding talking to people because my emotions are unpredictable and I worry that the mask might slip.''}
\end{tcolorbox}
\label{fig:example_instruction_gpt4}
\end{figure}

\begin{figure*}[!th]
\centering
\small
\caption{\textbf{Example Output of GPT-4.} The output from \gptfour\ shows the assessment results in a nested JSON format, following the steps outlined in the instruction. 
\hlc[Brown!15]{Step 1} contains \phq symptoms identified by \gptfour to be explicitly mentioned in the depression essay, along with the indicative excerpts and a symptom severity score. 
\hlc[Lavender!50]{Step 2} has all the remaining \phq symptoms accompanied by a free form reason text and an estimated severity score.
Finally, \hlc[gray!10]{Step 3} has the combined estimated \phq score and an overall degree of depression category. 
}
\begin{tcolorbox}[
    width=\textwidth,
    colback=white,
    colframe=black,
    arc=4mm,
    boxrule=0.5pt,
    left=2mm,
    right=2mm,
    top=2mm,
    bottom=2mm,
    fonttitle=\bfseries,
    ]


\begin{tcolorbox}[
    colback=Brown!15,
    boxrule=0pt,
    colframe=white,
    left=0mm,
    right=0mm,
    top=0mm,
    bottom=0mm,
    ]
\textbf{STEP 1:} \{

\quad Depressed Mood: [`I feel overwhelmed', 2],

\quad Worthlessness or Guilt: [`annoyed with myself for allowing it to continue', 2],

\quad Difficulty Concentrating: [`my emotions are unpredictable', 0]

\}
\end{tcolorbox}

\begin{tcolorbox}[
    colback=Lavender!50,
    boxrule=0pt,
    colframe=white,
    left=0mm,
    right=0mm,
    top=0mm,
    bottom=0mm,
    ]
\textbf{STEP 2:} \{

\quad Anhedonia: [``Based on the overall tone of hopelessness and resentment, it is likely that the author is experiencing diminished interest or pleasure", 1],

\quad Insomnia or Hypersomnia: [``No explicit mention, but stress and emotional unpredictability may affect sleep patterns", 1],

\quad Fatigue: [``Feeling overwhelmed could imply a state of mental fatigue", 1],

\quad Poor appetite or overeating: [``No explicit mention, but stress often affects eating habits", 1],

\quad Psychomotor Agitation or Retardation: [```avoiding talking to people' may suggest a withdrawal that could be associated with psychomotor retardation", 1],

\quad Suicidal Ideation: [``No explicit mention of suicidal thoughts or self-harm", 0]

\}
\end{tcolorbox}


\begin{tcolorbox}[
    colback=gray!10,
    boxrule=0pt,
    colframe=white,
    left=0mm,
    right=0mm,
    top=0mm,
    bottom=0mm,
    ]
\textbf{STEP 3:} \{

\quad Final: [Combined Score: 9, Category: Moderate]

\}
\end{tcolorbox}

\end{tcolorbox}
\label{fig:example_output_gpt4}
\end{figure*}

\subsection*{Machine Behavior Analysis of GPT-5}

\begin{figure}[!t]
\centering
\caption{\textbf{Similarities and differences between \gptfive's and self-report's schema of depressive symptoms.}}

\subfloat[\textbf{Pearson correlation between symptoms estimated by \gptfive     (left) and self-reported by participants (right).} The average correlations of each symptom with rest are presented as column vectors for both \gptfive and self-report. N=955. ]{
    \includegraphics[width=\linewidth]{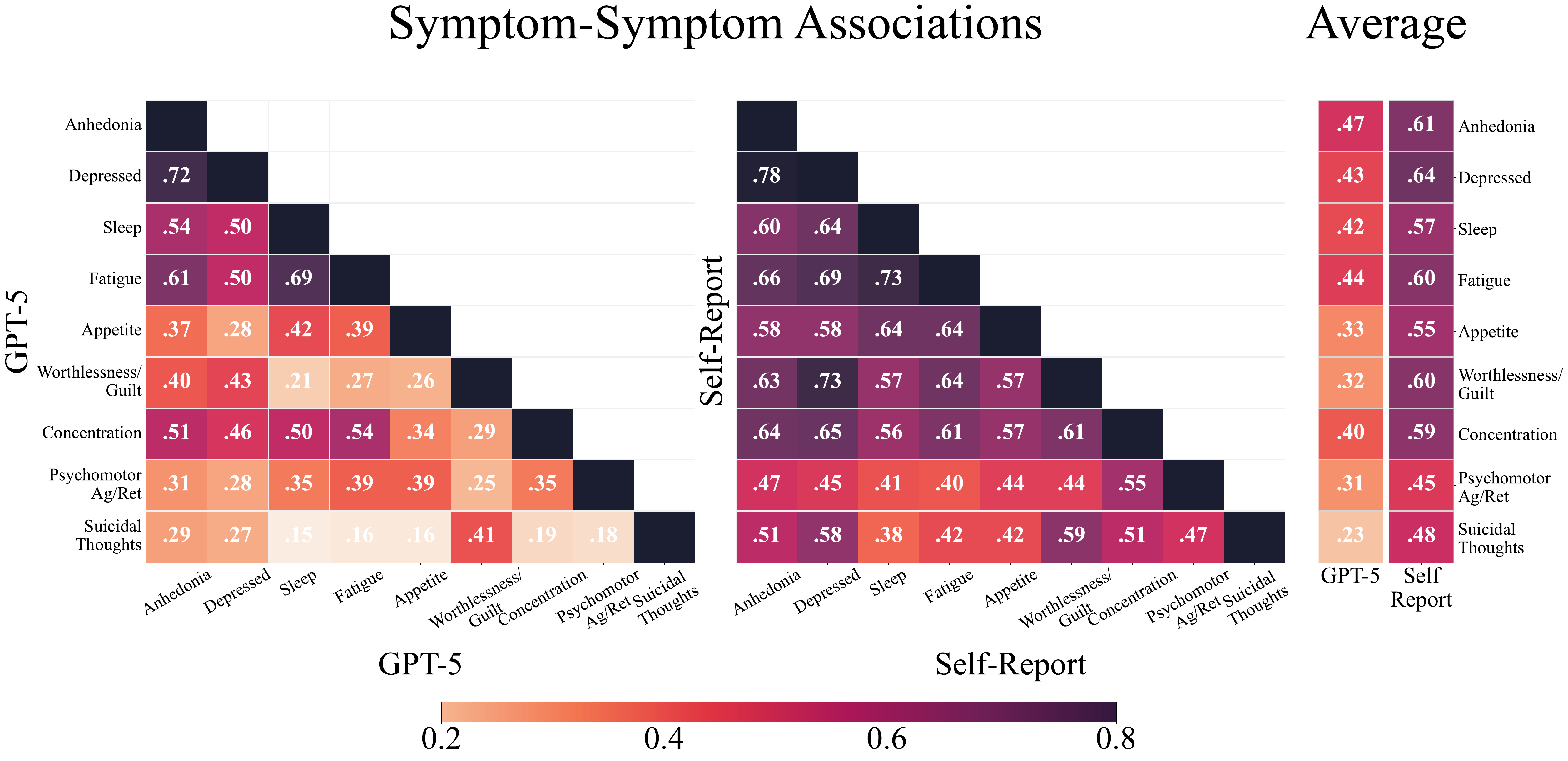}
    \label{fig:gpt5_sr_item_corr}
}
\par  
\subfloat[\textbf{Difference between \gptfive's and self-reports' Symptom-Symptom correlations.} 95\% confidence interval of the difference in correlations from bootstrapped resampling over 500 trials is enclosed within brackets. \gptfive's symptom-symptom relationship is weaker than self-reported.]{
    \includegraphics[width=0.50\linewidth]{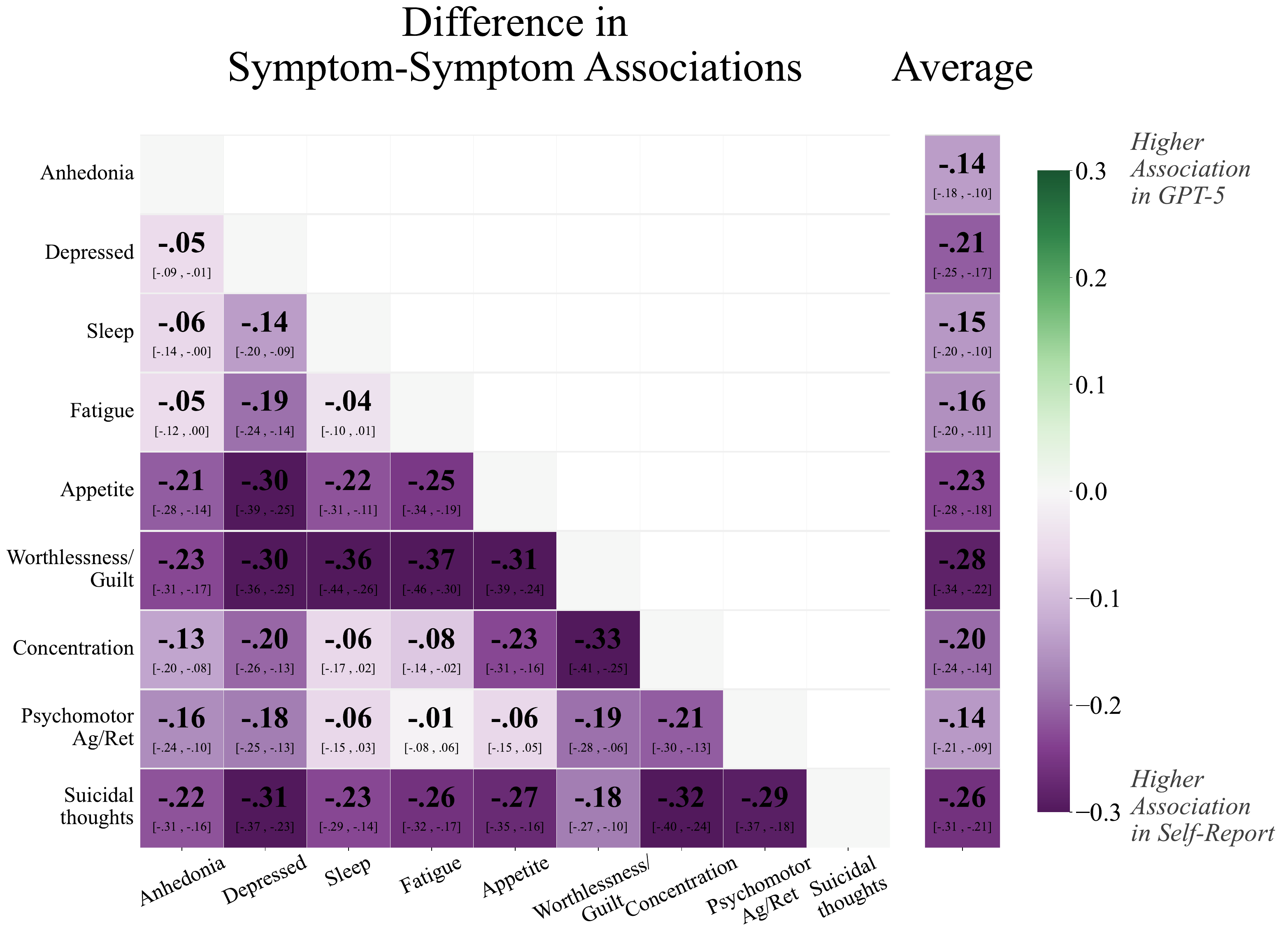}
    \label{fig:diff_gpt5_sr_item_corr}
}
\hspace{2pt}
\subfloat[\textbf{Symptom's item location ranking according to \textit{IRT}} over \gptfive and self-report scores for zero and non-zero decision boundary. 
    A lower rank value corresponds to the symptom being best at distinguishing lower values of depression, while a higher rank indicates being better at distinguishing higher values of depression. 
    Ranks are accompanied by a 95\% confidence interval estimated using bootstrapped resampling across individuals (N=955, num. trials=500).]{
    \begin{minipage}{0.45\linewidth}
    \fontsize{10pt}{9pt}\selectfont
    \begin{tabular}{l|P{1cm}P{1cm}P{1cm}}
    \hline
    \textbf{PHQ9 Item}   & \multicolumn{3}{c}{\textbf{Item Location Rank}}      \\
           & \textbf{\gptfive}    & \textbf{Self-Report}  & \textbf{Item location diff}  \\
    \hline
    Anhedonia & \cellcolor{rank2} 2 \tiny{[2, 2]} & \cellcolor{rank3} 3 \tiny{[3, 3]} & -1 \\ 
    Depressed Mood & \cellcolor{rank1} 1 \tiny{[1, 1]} & \cellcolor{rank2} 2 \tiny{[1, 2]} & -1 \\ 
    Sleep & \cellcolor{rank6} 6 \tiny{[4, 6]} & \cellcolor{rank4} 4 \tiny{[4, 5]} & 2 \\ 
    Fatigue & \cellcolor{rank3} 3 \tiny{[3, 3]} & \cellcolor{rank1} 1 \tiny{[1, 2]} & 2 \\ 
    Appetite & \cellcolor{rank9} 9 \tiny{[8, 9]} & \cellcolor{rank7} 7 \tiny{[7, 7]} & 2 \\ 
    Worthlessness / Guilt & \cellcolor{rank4} 4 \tiny{[4, 6]} & \cellcolor{rank5} 5 \tiny{[4, 6]} &  -1 \\ 
    Difficulty concentrating & \cellcolor{rank5} 5 \tiny{[4, 6]} & \cellcolor{rank6} 6 \tiny{[4, 6]} & -1 \\ 
    Psychomotor & \cellcolor{rank8} 8 \tiny{[8, 9]} & \cellcolor{rank8} 8 \tiny{[8, 8]} & 0 \\ 
    Suicidal Ideation & \cellcolor{rank7} 7 \tiny{[7, 7.5]} & \cellcolor{rank9} 9 \tiny{[9, 9]} & 0 \\ 
    \hline
    \end{tabular}
    \label{tab:irt_gpt5}
    \end{minipage}
}
\label{fig:gpt5_schema}
\end{figure}

The decrease in correlations of \gptfive assessments in comparison to \gptfour (see \tableautorefname~\ref{tab:item_item_corr}) was explained by its Schema of depression described below. \figureautorefname~\ref{fig:gpt5_schema} shows consistently weaker intersymptom relationship compared to self-report. 
The item location analysis also shows larger differences in discriminative properties of symptoms in \gptfive (\figureautorefname~\ref{tab:irt_gpt5}). 

\gptfive had a slightly higher rate of symptom identification (\tableautorefname~\ref{tab:exp_imp_convergence_gpt5}) than \gptfour for the less observed symptoms (Suicidal Ideation, Psychomotor agitation/retardation, Difficulty Concentration and Appetite).
However, its overall convergence with self-report was under both explicit and implicit identification showed a small decrease. 

\begin{figure}[!ht]
\centering
\caption{\textbf{Contrast of explicit \& implicit symptom mentions and convergence for \gptfive.} Pearson Correlation ($r$) between \gptfive and self-reported scores for all symptoms under conditions where symptoms were identified to be explicit and implicit (N=955) by \gptfive. 
    The bar chart in \% Explicit column represents the rate of explicit mentions according to \gptfive. 
    Higher correlations were observed consistently when symptoms were identified to be explicitly mentioned.}


{
    \begin{minipage}{\textwidth}
    \centering
    \fontsize{9pt}{4pt}\selectfont 
\begin{tabular}{m{3.5cm} >{\centering\arraybackslash}m{1.5cm} >{\centering\arraybackslash}m{5cm} >{\centering\arraybackslash}m{1.5cm}}
        \toprule
        \textbf{Symptom} & \textbf{Explicit $r$} & \textbf{\% Explicit vs. \% Implicit}  & \textbf{Implicit $r$} \\
        \midrule
        \vspace*{3pt}Anhedonia & \vspace*{3pt}.72 &  
        \begin{tikzpicture}[baseline=(current bounding box.center)]
        \begin{axis}[
            width=5cm, height=1.85cm, 
            axis lines=none, 
            xbar stacked,    
            bar width=0.4cm,   
            xmin=0, xmax=100, 
            ymin=0, ymax=1,   
            xtick=\empty,
            ytick=\empty,    
            clip=false, 
            ]
            \addplot [fill=barcolor] coordinates {(41.21,0)}; 
            \addplot [fill=cbarcolor] coordinates {(58.79,0)}; 
            \node[anchor=east, inner sep=0pt] at (axis cs:-2,0) {\makebox[15mm][r]{41.21\%}};
        \end{axis}
        \end{tikzpicture} & \vspace*{3pt}.52  \\
        \vspace*{3pt}Depressed Mood & \vspace*{3pt}.74 &  
        \begin{tikzpicture}[baseline=(current bounding box.center)]
        \begin{axis}[
            width=5cm, height=1.85cm,
            axis lines=none,
            xbar stacked,
            bar width=0.4cm,
            xmin=0, xmax=100,
            ymin=0, ymax=1,
            xtick=\empty,
            ytick=\empty,
            clip=false,
            ]
            \addplot [fill=barcolor] coordinates {(88.81,0)}; 
            \addplot [fill=cbarcolor] coordinates {(11.19,0)}; 
            \node[anchor=east, inner sep=0pt] at (axis cs:-2,0) {\makebox[15mm][r]{88.81\%}};
        \end{axis}
        \end{tikzpicture} & \vspace*{3pt}.51 \\
        \vspace*{3pt}Sleep & \vspace*{3pt}.53 & 
        \begin{tikzpicture}[baseline=(current bounding box.center)]
        \begin{axis}[
            width=5cm, height=1.85cm,
            axis lines=none,
            xbar stacked,
            bar width=0.4cm,
            xmin=0, xmax=100,
            ymin=0, ymax=1,
            xtick=\empty,
            ytick=\empty,
            clip=false,
            ]
            \addplot [fill=barcolor] coordinates {(12.45,0)}; 
            \addplot [fill=cbarcolor] coordinates {(87.55,0)}; 
            \node[anchor=east, inner sep=0pt] at (axis cs:-2,0) {\makebox[15mm][r]{12.45\%}};
        \end{axis}
        \end{tikzpicture} & \vspace*{3pt}.35 \\
        \vspace*{3pt}Fatigue & \vspace*{3pt}.57 & 
        \begin{tikzpicture}[baseline=(current bounding box.center)]
        \begin{axis}[
            width=5cm, height=1.85cm,
            axis lines=none,
            xbar stacked,
            bar width=0.4cm,
            xmin=0, xmax=100,
            ymin=0, ymax=1,
            xtick=\empty,
            ytick=\empty,
            clip=false,
            ]
            \addplot [fill=barcolor] coordinates {(17.36,0)}; 
            \addplot [fill=cbarcolor] coordinates {(82.64,0)}; 
            \node[anchor=east, inner sep=0pt] at (axis cs:-2,0) {\makebox[15mm][r]{17.36\%}};
        \end{axis}
        \end{tikzpicture} & \vspace*{3pt}.40 \\
        \vspace*{3pt}Appetite & \vspace*{3pt}.70 & 
        \begin{tikzpicture}[baseline=(current bounding box.center)]
        \begin{axis}[
            width=5cm, height=1.85cm,
            axis lines=none,
            xbar stacked,
            bar width=0.4cm,
            xmin=0, xmax=100,
            ymin=0, ymax=1,
            xtick=\empty,
            ytick=\empty,
            clip=false,
            ]
            \addplot [fill=barcolor] coordinates {(5.65,0)}; 
            \addplot [fill=cbarcolor] coordinates {(94.35,0)}; 
            \node[anchor=east, inner sep=0pt] at (axis cs:-2,0) {\makebox[15mm][r]{5.65\%}};
        \end{axis}
        \end{tikzpicture} & \vspace*{3pt}.20 \\
        \vspace*{3pt}Worthlessness / Guilt & \vspace*{3pt}.56 & 
        \begin{tikzpicture}[baseline=(current bounding box.center)]
        \begin{axis}[
            width=5cm, height=1.85cm,
            axis lines=none,
            xbar stacked,
            bar width=0.4cm,
            xmin=0, xmax=100,
            ymin=0, ymax=1,
            xtick=\empty,
            ytick=\empty,
            clip=false,
            ]
            \addplot [fill=barcolor] coordinates {(19.56,0)}; 
            \addplot [fill=cbarcolor] coordinates {(80.44,0)}; 
            \node[anchor=east, inner sep=0pt] at (axis cs:-2,0) {\makebox[15mm][r]{19.56\%}};
        \end{axis}
        \end{tikzpicture} & \vspace*{3pt}.28 \\
        \vspace*{3pt}Concentration & \vspace*{3pt}.46 & 
        \begin{tikzpicture}[baseline=(current bounding box.center)]
        \begin{axis}[
            width=5cm, height=1.85cm,
            axis lines=none,
            xbar stacked,
            bar width=0.4cm,
            xmin=0, xmax=100,
            ymin=0, ymax=1,
            xtick=\empty,
            ytick=\empty,
            clip=false,
            ]
            \addplot [fill=barcolor] coordinates {(11.82,0)}; 
            \addplot [fill=cbarcolor] coordinates {(88.18,0)}; 
            \node[anchor=east, inner sep=0pt] at (axis cs:-2,0) {\makebox[15mm][r]{11.82\%}};
        \end{axis}
        \end{tikzpicture} & \vspace*{3pt}.31  \\
        \vspace*{3pt}Psychomotor & \vspace*{3pt}.04 & 
        \begin{tikzpicture}[baseline=(current bounding box.center)]
        \begin{axis}[
            width=5cm, height=1.85cm,
            axis lines=none,
            xbar stacked,
            bar width=0.4cm,
            xmin=0, xmax=100,
            ymin=0, ymax=1,
            xtick=\empty,
            ytick=\empty,
            clip=false,
            ]
            \addplot [fill=barcolor] coordinates {(4.60,0)}; 
            \addplot [fill=cbarcolor] coordinates {(95.40,0)}; 
            \node[anchor=east, inner sep=0pt] at (axis cs:-2,0) {\makebox[15mm][r]{4.60\%}};
        \end{axis}
        \end{tikzpicture} & \vspace*{3pt}.11  \\ 
        \vspace*{3pt}Suicidal Ideation & \vspace*{3pt}.70 & 
        \begin{tikzpicture}[baseline=(current bounding box.center)]
        \begin{axis}[
            width=5cm, height=1.6cm,
            axis lines=none,
            xbar stacked,
            bar width=0.4cm,
            xmin=0, xmax=100,
            ymin=0, ymax=1,
            xtick=\empty,
            ytick=\empty,
            clip=false,
            ]
            \addplot [fill=barcolor] coordinates {(12.97,0)}; 
            \addplot [fill=cbarcolor] coordinates {(87.03,0)}; 
            \node[anchor=east, inner sep=0pt] at (axis cs:-2,0) {\makebox[15mm][r]{12.97\%}};
        \end{axis}
        \end{tikzpicture} & \vspace*{3pt}.05 \\
        \bottomrule
        \vspace*{3pt}\textbf{Mean} & \vspace*{3pt}.56 & 
        \begin{tikzpicture}[baseline=(current bounding box.center)]
        \begin{axis}[
            width=5cm, height=1.6cm,
            axis lines=none,
            xbar stacked,
            bar width=0.4cm,
            xmin=0, xmax=100,
            ymin=0, ymax=1,
            xtick=\empty,
            ytick=\empty,
            clip=false,
            ]
            \addplot [fill=barcolor] coordinates {(23.82,0)}; 
            \addplot [fill=cbarcolor] coordinates {(76.18,0)}; 
            \node[anchor=east, inner sep=0pt] at (axis cs:-2,0) {\makebox[15mm][r]{23.82\%}};
        \end{axis}
        \end{tikzpicture} & \vspace*{3pt}.30 \\
        \bottomrule
    \end{tabular}
    \label{tab:exp_imp_convergence_gpt5}
    \end{minipage}
}

\end{figure}

\end{document}